\documentclass[preprint,12pt]{elsarticle}

\usepackage{subfigure}  % to better equalize the last page
\usepackage{url}      % llt: nicely formatted URLs

\usepackage{algorithm}
\usepackage{algpseudocode,multirow}
\usepackage{amsmath}
\usepackage{amssymb}
\usepackage[super]{nth}
\usepackage{graphicx,amsmath,url}
\DeclareMathOperator*{\argmax}{arg\,max}

\newcommand{\commentout}[1]{}
\begin{document}
\begin{frontmatter}

  \title{A Difficulty Ranking Approach to Personalization in E-learning}
  
% AUTHORS
\author[bgu]{Avi Segal \corref{cor1}} 
\ead{avise@post.bgu.ac.il}

\author[bgu]{Kobi Gal}
\ead{kobig@bgu.ac.il}

\author[bgu]{Guy Shani}
\ead{shanigu@bgu.ac.il}

\author[bgu]{Bracha Shapira}
\ead{bracha.shapira@gmail.com}

\cortext[cor1]{Corresponding author.}

% ADRESSES

\address[bgu]{Ben-Gurion University of the Negev, Israel}

\begin{abstract} The  prevalence of e-learning systems and on-line
  courses has made educational material widely accessible to students
  of varying abilities and backgrounds.  There is thus a
  growing need to accommodate for individual differences in e-learning
  systems.  This paper presents an algorithm called EduRank for
  personalizing educational content to students that combines a
  collaborative filtering algorithm with voting methods.
  EduRank constructs a difficulty ranking for each student by
  aggregating the rankings of similar students using different aspects
  of their performance on common  questions.  These aspects
  include grades, number of retries, and time spent solving questions.
  It infers a difficulty ranking directly over the questions for each
  student, rather than ordering them according to the student's
  predicted score.  The EduRank algorithm was tested on two data sets
  containing  thousands of students and a million records.  It
  was able to  outperform the state-of-the-art ranking
  approaches as well as a domain expert.  EduRank was used by students in 
   a classroom activity, where a prior model was incorporated to predict the difficulty rankings of students with  no prior history in the system. It was shown to lead students to
  solve more difficult questions than an ordering by a domain expert,
  without reducing their performance.
 % We also show that incorporating a prior
 %  distribution over students' difficulty rankings can alleviate the
 %  cold-start problem that arises when the algorithm is used in a
 %  classroom setting.
  % Our approach can potentially be used to support
  % teachers in tailoring problem sets and exams to individual students
  % and students in informing them about areas they may need to
  % strengthen.
\end{abstract}

\begin{keyword}
  Artificial Intelligence \sep e-learning \sep  ranking algorithms \sep
  human-computer decision-making
%% keywords here, in the form: keyword \sep keyword

%% PACS codes here, in the form: \PACS code \sep code

%% MSC codes here, in the form: \MSC code \sep code
%% or \MSC[2008] code \sep code (2000 is the default)

\end{keyword}

\end{frontmatter}

%\maketitle

\section{Introduction}

The prevalence of educational software in schools and the explosion of
on-line course opportunities have made educational content a common
resource that is accessible to student communities of varied
backgrounds and  learning abilities.  There is thus a growing need
for personalizing educational content to students in e-learning
systems in a way that adapts to students' individual
needs~\cite{sampson2010personalised,akbulut2012adaptive,ba2007framework,zhang2008personalized,najar2014adaptive,mazziotti2015robust}.

Many educational applications present a sequence of questions to students, ordered by increasing difficulty. The student is expected to first solve easier questions in a given skill, and only after mastering the skill, move to more difficult and challenging questions. As such, ordering questions by difficulty is an important task in such applications.

This paper provides a novel approach for personalization of educational content  that directly creates a {\em difficulty ranking} over new questions allowing us to order questions differently for different students. 
Our approach is based on collaborative filtering~\cite{breese1998empirical}, 
which is a commonly used technique  in recommendation systems for predicting the interests of a user by collecting preferences from their online activities.
The explosive growth of e-commerce and online environments mean that users are becoming overloaded by options to consider and they may not have the time or knowledge to personally evaluate these options. Recommender systems  have proven to be a valuable way for online users to cope with the information overload and have become one of the most powerful and popular tools in electronic commerce~\cite{shani2011evaluating}.

This paper uses the collaborative filtering approach to generate  a difficulty ranking over a set of questions for a target student by aggregating the known difficulty rankings over questions solved by other, similar students.  The similarity of other students to the target student is measured by their grades on common past questions, the number of retries for each question, and other features. Unlike other applications of collaborative filtering in education, our approach directly generates a difficulty ranking over the test questions, avoiding the need to predict the students' performance directly on these questions, which is prone to error. For example, in the KDD cup 2010, the best performing grade prediction algorithms exhibited an error rate of about $28\%$~\cite{toscher2010collaborative}. We demonstrate the same problem over the datasets that we used in our empirical analysis.

Our algorithm, called EduRank, weighs the contribution of these
students using measures from the information retrieval literature. It
allows for partial overlap between the difficulty rankings of a
neighboring student and the target student, making it especially
suitable for e-learning systems where students differ on which
questions they solve.  The algorithm extends a prior approach for
ranking items in recommendation systems~\cite{liu2008eigenrank}, which
was not evaluated on educational data, in two ways: First, by using voting methods from 
social choice~\cite{brandt2012computational} to combine the difficulty rankings of similar
students and produce a better difficulty ranking for the target
student. Second, EduRank penalizes disagreements in high positions in
the difficulty ranking more strongly than low positions, under the
assumption that errors made when ranking more difficult questions are
more detrimental to students than errors made when ranking easier
questions.  EduRank can support both teachers and students by
automatically tailoring problem sets or exams to the abilities of
individual students in the classroom, or by informing students about
topics which they need to strengthen.

We evaluated EduRank on two large  data sets containing tens of thousands of students and about a million records. We compared the performance of EduRank to a variety of personalization methods from the literature, focusing on popular collaborative filtering approaches such as matrix factorization and memory-based nearest neighbours. We also compared EduRank to a (non-personalized) ranking created by a domain expert. EduRank was able to  outperform all other approaches when comparing the outputted difficulty rankings to a gold standard. 

 EduRank was embedded  in a real classroom and used  to sequence math questions to students by inferred order of difficulty.  Its performance was compared to an alternative sequencing approach that  selected questions by increasing order of difficulty, as determined by pedagogical experts. We found that students using the EduRank 
algorithm solved harder questions and spent more time in the system than the   expert-based sequencing approach, without impeding their overall performance. This demonstrates the potential of the EduRank approach to contribute 
to students' learning in the classroom.

The contributions of this paper are three-fold. First, it  presents a novel algorithm for personalization in e-learning according to the level of difficulty by combining collaborative filtering with social choice. Second, it is shown to  outperform alternative ranking solutions from the literature on two real-world data sets. Finally, it is shown to improve students' performance in the classroom when solving questions of different difficulty levels. 
 This paper extends prior work describing the EduRank
algorithm~\cite{EduRank} in several ways.  First, we show  there is low
agreement  in the difficulty rankings between students over the same set of questions. This
demonstrates that sequencing questions to student in a ``one-size-fits-all" approach cannot address these
individual differences among students.  Second, we
have extended the algorithm to handle the ``cold-start problem'' in
e-learning systems in which there is a need to sequence material for
new students with little or no history in the system.  Third, we 
deployed the algorithm in a real classroom, where it was compared to an alternative sequencing 
approach that was designed by pedagogical experts. 
%
%include a comparison with an additional kind of collaborative
%filtering approach which is based on Maximum Margin Matrix
%Factorization.}
 
% The rest of this paper is organized as follows:
% Section~\ref{sec:background} provides the definitions of basic
% measures from information retrieval that are used later in the
% paper. Section~\ref{sec:ProblemDefinition} defines the difficulty
% ranking problem and presents the EduRank algorithm for solving this
% problem. Section~\ref{sec:empirical} compares EduRank to alternative
% approaches on data sets of students' interactions with e-learning
% software. Section~\ref{sec:class} describes how EduRank was configured
% to sequence questions to students and deployed in a real classroom,
% where it was compared to an expert-based sequencing approach.
% Section~\ref{sec:related} provides related work from the learning
% technologies and recommendation systems literature.  Lastly,
% Section~\ref{sec:conc} concludes the paper and describes future work.

\section{Background}
\label{sec:background}
In this section we briefly review relevant approaches and metrics in recommendation systems, social choice, and information retrieval.

\subsection{Online Learning and Intelligent Tutoring Systems}

Intelligent Tutoring Systems have been used for computer based instruction since the 1970s. Seeking to apply artificial intelligence techniques for ``intelligent'' computer-based instruction, their goal is to engage students in sustained reasoning activities and to interact with the student while understanding the student behavior and state.   Graesser et. al \cite{graesser2012intelligent} reviewed the state of ITS and specifically the research on different classes of ITS. They describe the computational mechanism of each type of ITS and the available empirical assessments of the impact of these systems on learning gains.

In recent years we have seen a dramatic change in the education world towards wide-adoption of online learning technologies (e-learning).  The huge amount of fine-grained data being collected, coupled with Big Data and artificial-intelligence mechanisms, can be used to develop learning environments that can adapt to the needs of the individual learner. For example, MOOCs have democratized the access to educational resources, making them accessible to anyone with an internet connection~\cite{kizilcec2017towards,clow2013moocs}. 

Due to the growing prevalence of online learning settings, how to sequence educational content to students  is an important research problem.  Existing work in this area focused on using graph search \cite{zhao2006shortest}, neural networks \cite{idris2009adaptive},  and heuristic semantics \cite{al2011heuristic} for learning path personalization, among other methods. 

\subsection{Recommendation Systems and Collaborative Filtering}
\label{sec:receng}
Recommender systems actively help users in identifying items of interest. The prediction of users' ratings for items, and the identification of the top-N relevant items to a user, are popular tasks in recommendation systems.  For example, a recommendation system for movies, may predict the rating that a user may give to a newly released movie, or recommend a new movie for a user given movies that the user liked in the past. A commonly used approach for both tasks is {\em collaborative filtering} (CF), which uses data over other users, such as their ratings, item preferences, or performance in order to compute a recommendation for the active user.

There are two common collaborative filtering approaches~\cite{breese1998empirical}. In the memory-based nearest neighbors approach, a similarity metric, such as the Pearson correlation, is used to identify a set of neighboring users. The predicted rating for a target user and a given item can then be computed using a weighted average of ratings of other users in the neighborhood. In the model-based approach, a statistical model between users and items is created from the input data. For example, the popular matrix factorization approach~\cite{sarwar,KorenB15} computes a latent feature vector for each user and item, such that the inner product of a user and item vectors is higher when the item is more appropriate for the user.

While rating prediction and top-N recommendations are widely researched, there are only a few attempts to use CF approaches to generate rankings. Of these, most methods order items for target users according to their predicted ratings.  In contrast, Liu et al. developed the EigenRank algorithm ~\cite{liu2008eigenrank} which is a CF approach that relies on the similarity between item ratings of different users to directly compute the recommended ranking over items.  They show this method to outperform existing collaborative filtering methods that are based on predicting users' ratings.  EigenRank computes the similarity of users using the Kendall $\tau$ metric --- a well known metric in information retrieval for comparing two rankings which counts the number of pairwise disagreements between the two lists --- rather than by metrics such as Pearson correlation which are popular in rating prediction. Matrix factorization methods have also been suggested for ranking~\cite{weimer2008cofi}. 

%The CofiRank algorithm \cite{WeimerKLS07,WeimerKS08} uses a maximum margin matrix factorization with a trace norm regulalarization, miminimizing either NDCG or squared loss. CofiRank was shown to perfrom very well on movie ranking tasks.

Using the ratings of similar users, EigenRank computes for each pair of items in the query test set so-called potential scores for the possible orderings of the pair. Afterward, EigenRank converts the pair-wise potentials into a ranked list. EigenRank was applied to movie recommendation tasks, and was shown to order movies by ranking better than methods based on converting rating predictions to a ranked list.

\subsection{Social Choice}

\label{sec:copeland}
{\em Social choice} theory originated in economics and political science,
and is dealing with the design and formal analysis of methods for
aggregating preferences (or votes) of multiple
agents~\cite{fishburn1973theory}. Examples of such methods include
voting systems used to aggregate preferences of voters over a set of
alternatives to determine which alternative(s) wins the election,
and systems in which voters rank a complete set of alternatives using an
ordinal scale.  One such approach which we use in this paper is
Copeland's method~\cite{copeland1951reasonable,nurmi1983voting}
ordering alternatives based on the number of pairwise defeats and
victories with other alternatives.

The Copeland score for an alternative $q_j$ is determined by taking the
number of those alternatives that $q_j$ defeats and subtracting from
this number those alternatives that beat $q_j$.  A partial order over
the items can then be inferred from these scores. Two advantages of
this method that make it especially amenable to e-learning systems
with many users (e.g., students and teachers), and large data sets, are
that they are quick to compute and easy to explain to
users~\cite{schalekamp2009rank}.  Pennock et
al.~\cite{pennock2000social} highlighted the relevance of social
choice to CF, demonstrating that properties and limitations from Social Choice theory apply in the context of CF. Following their work, we consider weighted versions of voting mechanisms to CF algorithms and demonstrate the effect of this approach to e-learning systems.

% More recently, different computer algorithms and methods have been proposed to study various aspects of social choice, like the complexity of different voting mechanisms and voting procedures for fair division.

% as summarized by!\cite{brandt2012computational} deals with the application of methods, namely computer science algorithms, to problems of social choice. Schalekamp and Zuylen~\cite{schalekamp2009rank} give a good overview of social choice algorithms for aggregating preferential votes of multiple users. Typically in these algorithms each voters is given the same voting power when aggregating votes.

\subsection{Metrics for Ranking Scoring}
\label{sec:metrics}
A common task in information retrieval is to order a list of results according to their relevance to a given query~\cite{zhou2010evaluating}.  Information retrieval methods are typically evaluated by comparing their proposed ranking to that of a gold standard, known as a {\em reference ranking}, which is provided by the user or by a domain expert, or inferred from the data.  

Before describing the comparison metrics and stating their relevance for e-learning systems, we define the following notations: Given a set $L$ of questions of varying difficulties, let $\binom{L}{2}$ denote the set of all non ordered pairs in $L$.  Let $\succ$ be a partial order over the set of questions $L$. We define the reverse order of $\succ$ over $L$, denoted $\overline{\succ}$ as a partial order over $L$ such that if $q_j \succ q_k$ then $q_k \overline{\succ} q_j$.  Let $\succ_1$ and $\succ_2$ be two partial orderings over a set of questions $L$, where $\succ_1$ is the reference order and $\succ_2$ is the system proposed order. We define 
an agreement relation between the orderings $\succ_1$ and $\succ_2$ as follows:
\begin{itemize}
\item  The orderings $\succ_1$ and $\succ_2$ \emph{agree} on questions $q_j$ and $q_k$ if $ q_j \succ_1 q_k$ and $q_j \succ_{2} q_k$. That is, both $\succ_1$ and $\succ_2$ order the two questions in the same order.
\item  The orderings $\succ_1$ and $\succ_2$ \emph{disagree} on questions $q_j$ and $q_k$ if $ q_j \succ_1 q_k$ and $q_k \succ_{2} q_j$. That is, while in $\succ_1$ $q_j$ is deemed more difficult than $q_k$, in $\succ_2$ $q_j$ is considered to be easier than $q_k$.
\item The orderings $\succ_1$ and $\succ_2$ \emph{are compatible} on questions $q_j$ and $q_k$ if $q_j\succ_1 q_k$ and neither $q_j\succ_2 q_k$ nor $q_k\succ_2 q_j$. That is, $\succ_2$ does not provide any ordering of questions that are ordered in $\succ_1$. This is not symetric --- if the reference order $\succ_1$ does not provide any ordering over a pair, any ordering provided by the system ordering $\succ_2$ is acceptable.
\end{itemize}
Given a partial order $\succ$ over questions $Q$, the restriction of $\succ$ over $L\subseteq Q$ are all questions $(q_k,q_l)$ such that $q_k \succ q_l$ and $q_k,q_l\in L$. That is, we restrict the order only to a subset of questions that are of interest to us, for example, questions that a specific student has already answered.

\subsubsection {Normalized Distance based Performance} 

The Normalized Distance based Performance Measure (NDPM)~\cite{yao1995measuring,shanievaluating} is a commonly used metric for evaluating a proposed system ranking to a reference ranking. It differentiates between correct orderings of pairs, incorrect orderings and ties.  Formally, let $\delta_{\succ_1, \succ_2} (q_j, q_k)$ be a distance function between a reference ranking $\succ_1$ and a proposed ranking $\succ_2$ defined as follows:

\begin{equation}
  \label{eq:dist}
\delta_{\succ_1, \succ_2} (q_j, q_k)=
\begin{cases}
0 & \text{if }  \succ_1 \text{and } \succ_2 \text {agree on } q_j \text{and } q_k,\\
1 & \text{if }  \succ_1 \text{and } \succ_2 \text {are compatible on } q_j \text{and } q_k,\\
2 & \text{if }  \succ_1 \text{and } \succ_2 \text {disagree on } q_j \text{and } q_k.\\
\end{cases}
\end{equation} 

% We let the function equals 0 if $\succ_1$ and $\succ_2$ agree on $q_j$ and $q_k$, 1 if $\succ_1$ and $\succ_2$ are compatible on $q_j$ and $q_k$, and 2 if $\succ_1$ and $\succ_2$ disagree on $q_j$ and $q_k$.
The total distance over all question pairs in $L$ is defined as follows \begin{equation}
  \label{eq:2}
\beta_{\succ_1, \succ_2}(L)=\sum_{(q_j,q_k)\in \binom{L}{2}} \delta_{\succ_1, \succ_2} (q_j, q_k)
\end{equation}

%Let $\succ^*(L)=\textnormal{argmin}_{\succ} \beta_{\succ,\succ_2}(L)$  be the ranking that maximizes the distance function $\delta(\succ_1,\succ_2)$.
%The NDPM score $\tau_{ND}$ is defined as 
%\begin{equation}
 % \label{eq:1}
 % s_{ND}(L,\succ_1,\succ_2)=\frac{\beta_{\succ_1,\succ_2}(L)}{\succ^*(L)}
%\end{equation}

Let $ m(\succ_1) = \argmax_{\succ}  \beta_{\succ_1,\succ}(L)$ be a normalization factor which is the maximal distance that any ranking $\succ$ can have from a reference ranking $\succ_1.$ The NDPM score $s_{ND}(L,\succ_1,\succ_2)$ comparing a proposed ranking of questions $\succ_2$  to a reference ranking $\succ_1$ 
is defined as \begin{equation}
  \label{eq:1}
  s_{ND}(L,\succ_1,\succ_2)=\frac{\beta_{\succ_1,\succ_2}(L)}{m(\succ_1)}
\end{equation}
Intuitively, the NDPM measure will give a perfect score of 0 to difficulty rankings over the set in $L$ that completely agree with the reference ranking, and a worst score of 1 to a ranking that completely disagrees with the reference ranking. If the proposed ranking does not contain a preference between a pair of questions that are ranked in the reference ranking, it is penalized by half as much as providing a contradicting preference.  

The evaluated ranking is not penalized for containing preferences that are not ordered in the reference ranking.  This means that for any question pair that were not ordered in the true difficulty ranking, any ordering predicted by the ranking algorithm is acceptable. Not penalizing unordered pairs is especially suitable for e-learning systems, as well as other collaborative filtering applications, in which many questions for the target student in $L$ may not have been solved by other students and these questions may remain unordered in the difficulty ranking.

\subsubsection {AP Rank Correlation}
A potential problem with the NDPM metric is that it does not consider the location of disagreements in the reference ranking.  In some cases it is more important to appropriately order items that should appear closer to the head of the ranked list, than items that are positioned near the bottom. For example, when ranking movies, it may be more important to properly order the  movies that the user would enjoy, than to properly order the movies that the user would not enjoy.  

Similarly, we assume that the severity of errors in ranking questions depends on their position in the ranked list.  As we are interested in sequencing questions by order of difficulty, properly predicting how easy questions should be ordered is not as important as avoiding the presentation of a difficult question too early, resulting in frustration and other negative effects on the student learning process.
Therefore, when evaluating a ranked list of questions, it is often important to consider the position of the questions in the ranked list. We would like to give different weights to errors depending on their position in the list.

To this end, we can use the AP correlation metric~\cite{yilmaz2008new}, which gives more weight to errors over items that appear 
at higher positions in the reference ranking.  % Average precision can be defined as the average
% of the precisions at relevant documents, where the precisions at
% unretrieved documents are assumed to be zero. Yilmaz, Aslam and
% Roberson \cite{yilmaz2008new} proposed a new rank correlation
% statistics based on the probabilistic interpretation of average
% precision that gives more importance to the items towards the top of
% the list.
Formally, let $\succ_1$ be the  reference ranking and $\succ_2$ be a
proposed ranking over a set of items.  The AP measure compares the order
between each item in the proposed ranking $\succ_2$ with all items that
precede it with the ranking in the reference ranking $\succ_1$.

For each $q_k, q_j \in L,  k \neq j$, let the set $Z^k(L,\succ_2)$ denote all question pairs $(q_k, q_j)$ in $L$ such that 
$q_j\succ_2 q_k$. These are all the questions that are more difficult to the student than question $q_k$.
\begin{equation}
  \label{eq:22}
  Z^k(L,\succ_2) = \{ (q_j, q_k) \mid  \forall q_j\neq q_k \textnormal{ s.t. }
  q_j \succ_2 q_k \textnormal{ and } q_j, q_k \in L \} 
\end{equation}
We define the indicator function $I^A(q_j, q_k,\succ_1,\succ_2)$ to equal
1 when $\succ_1$ and $\succ_2$ {agree} on questions $q_j$ and
$q_k$. 

Let $A^k (L,\succ_1,\succ_2)$ be the normalized agreement score between $\succ_2$ and the reference ranking $\succ_1$ for all questions $q_j$ such that 
$q_j \succ_{i} q_k$. 
\begin{equation}
  \label{eq:3}
  A^k(L,\succ_1,\succ_2) = \frac{1}{k-1}
\sum\limits_{(q_j,q_k) \in Z^k(L,\succ_2) } I^A(q_j,q_k,\succ_1,\succ_2) 
\end{equation}
The AP score of a partial order $\succ_2$ over $L$ given partial order
$\succ_1$ is
defined as 
\begin{equation}
  \label{eq:4}
  s_{AP}(L,\succ_1,\succ_2) = \frac{1}{|L|-1} \sum_{k=2}^{|L|} A^k(L,\succ_1,\succ_2)
\end{equation}
The
 $ s_{AP}$ score gives a perfect score of 1 to systems where there is total agreement between the system proposed difficulty ranking and the reference ranking for every question pair  above location $i$ for all $i \in 
 \{ 1, \ldots, |L| \}$. The worst score of 0 is given to systems were there is no agreement between the two ranked lists.

\section{Problem Definition and the EduRank Algorithm}
\label{sec:ProblemDefinition}

 The  {\em difficulty ranking} problem is defined by a target student $s_i$, and a set of questions $L_i$, for which the algorithm must predict a difficulty ranking $\hat{\succ_i}$. The predicted difficulty ranking $\hat{\succ_i}$ is evaluated with respect to a difficulty reference ranking $\succ_i$ over $L_i$ using a scoring function $s(\hat{\succ_i}, \succ_i,L_i)$.  

To solve this problem, we take a collaborative filtering approach, which uses the difficulty rankings on $L_i$ of other students similar to $s_i$ to construct a difficulty ranking over $L_i$ for student $s_i$.  Specifically, the input to the problem is:
\begin{enumerate}
\item A set of students $S = \{s_1, s_2, ..., s_m\}$.
\item A set of questions $Q = \{q_1, q_2, ..., q_n\}$.
\item For each student $s_j \in S$, a partial difficulty ranking $\succ_j$ over a set of questions $T_j \subseteq Q$.
\item For each student $s_j \in S$, a subset of questions $L_j \subseteq Q$ that must be ordered.
\end{enumerate}

For every student $s_j\in S$ we require two disjoint subsets  $T_j, L_j \in Q$, where the difficulty ranking of $s_j$ over $T_j$ is known, and is a restriction of $\succ_j$ over all the questions in $Q$. Intuitively, for a a target student $s_i\in S$,  $T_i$ represent the set of questions that the target student $s_i$ has already answered, while $L_i$ is the set of questions for which a difficulty ranking 
needs to be predicted.  For example, $L_i$ may be a set of questions in a homework assignment, that needs to be ordered for a particular student, while $T_i$ may be the set of all questions that the student has solved prior to this homework assignment.

The collaborative filtering task is to leverage the known rankings of all students $s_j$ over $T_j$ in order to compute the required difficulty ranking $\hat{\succ_i}$ over $L_i$ for student $s_i$.  % Obvious alternatives for computing the ranking comparison score are the NDPM and AP scoring metrics that we review above.

We now present our EduRank algorithm for producing a personalized
difficulty ranking over a given set of questions $L_i$ for a target
student $s_i$. EduRank estimates how similar other students are to $s_i$,
and then combines the ranking of the similar students over $L_i$ to
create a ranking for $s_i$.  There are two main procedures to the
algorithm: computing the student similarity metric, and
creating a difficulty ranking based on the ranking of similar users. 

For comparing the target student $s_i$ to potential neighbors, we use the
$s_{AP}$ metric over questions in $T_i$. We prefer $s_{AP}$ to, e.g., NDPM, to encourage greater similarity between students with high agreement
in top positions (more difficult questions) in their respective rankings.  

For aggregating the different students' rankings to create a
difficulty ranking for the target student, we use the Copeland method.
We treat each question as an alternative and  look at the aggregated voting of neighbors based on their similarity metric. In our aggregated voting calculation, alternative $i$ is preferred over alternative $j$ if the similarity normalized number of wins of $i$ over $j$ computed over all neighbors is higher than the similarity normalized number of losses. The Copeland method then computes for each alternative question the overall number
of aggregated victories and aggregated defeats and ranks the alternatives accordingly.  Let the win score $\gamma(q_k,q_l,\succ)$ over question pairs $q_k,q_l$ given a
difficulty ranking $\succ$ as follows:
\begin{equation}
  \label{eq:a2}
  \gamma(q_k,q_l,\succ) = 
\begin{cases}
1 & \text{if } q_k \succ q_l \\
-1 & \text{if } q_l \succ q_k\\
0 & \text{otherwise }
\end{cases}
\end{equation}

The relative voting  $rv(q_k, q_l,S)$ of two questions $q_k, q_l$ given the difficulty
rankings of a group of (neighboring) students $S$ is
\begin{equation}
  \label{eq:a9}
 rv(q_k, q_l,S) =  sign(\sum_{j \in S\setminus i} s_{AP}(T_i,\succ_i,\succ_j) \cdot  \gamma(q_k,
  q_l,\succ_j))
\end{equation}

The Copeland score $c(q,S,L_i)$ of a question $q$ given the difficulty
rankings of a set of students S  and a subset of questions $L_i$ is  
\begin{equation}
  \label{eq:a1}
  c(q,S,L_i) = \sum_{q_l\in L_i\setminus q} rv(q, q_l,S)
\end{equation}

% The Copeland score of question $q$ can be measured as the difference
% between the number of
% questions in $L_i$ that are less preferred to $q$ and the number of
% questions in $L_i$ that are more preferred to $q$ according to the
% partial order $\succ_j$ of student $s_j$.  

The EduRank algorithm is shown in Figure~\ref{alg:edurank}. The
input to the EduRank algorithm is a set of students
$S=\{s_1,\ldots,s_n\}$, each with a known ranking over a set of
questions $T_j$, such that $Q= T_1 \cup\ldots  \cup T_n$. In addition the
algorithm is given a target student $s_i \in S$, and a set of
questions $L_i \subseteq Q$ that needs to be ranked for $s_i$. The
output of the algorithm is a ranking of the questions in $L_i$.

The algorithm computes a ranking score $c(q)$ for each question
$q \in L_i$, which is the Copeland score for that question, as defined above.
 The algorithm returns a partial order for student $s_i$ over the test set
$L_i$ where questions are ranked by decreasing Copeland score
$c(q)$.

\par{Complexity: }The bottleneck of the computational complexity of the algorithm is computing the relative voting score  in line 3 of Figure~\ref{alg:edurank}. As shown in Equation~\ref{eq:a9} the relative voting of  a question pair $(q_k,q_l)$ for student $s_i$ combines the similarity 
score $s_{AP}$ with the ranking score $\gamma$.  To compute the similarity score between $s_i$ and each student in $S$, 
all  students in $S$ are traversed once, and for each student all pairs of questions in $L_i$ are traversed. Thus, the similarity computation complexity is $\mathcal{O}(S\cdot|L_i|^2)$. To compute the ranking score all of $s_i$ neighbors are traversed once, and for each neighbor all question pairs are compared for each question location. This leads to a ranking computation complexity bounded by $\mathcal{O}(S\cdot|L_i|^4)$. This complexity is polynomial in the number of questions in the test set of the target student. 
% compute the similarity between complexity of this 
% operation is computation complexity for generating a ranked list for a given student $s_i$ consists of the similarity computation complexity and the ranking computation complexity. For similarity computation, all students in $S$ are traversed once, and for each student all questions in $L_i$ are traversed once. Thus, the similarity computation complexity is $\mathcal{O}(S\cdot|L_i|)$. For  ranking computation, all of $s_i$ neighbors are traversed once, and for each neighbor all all question pairs are compared. Thus, the ranking computation complexity is bounded by $\mathcal{O}(S\cdot|L_i|^2)$

% Note that as a collaborative filtering approach, EduRank allows for
% partial overlap between the difficulty rankings of a neighboring
% student and the target student. That is, EduRank combines the rankings
% of students who answered only some of the questions that the target
% student $s_i$ has answered.  

%\commentout{
%Second,
%the computational complexity of this algorithm (for each target
%student) is polynomial in the number of questions in the
%database. Intuitively, for each question in the test set $L_i$, the
%Copeland score can be computed in linear time in the size of the test
%set.  Assuming the number of neighboring students can be computed offline, the
%similarity measure can be computed in quadratic time in the size of
%the training set $T_i$.
%}

\begin{figure}[t]
\textbf{INPUT:}
Set of students $S$; 
Set of questions $Q$; 
For each student $s_j \in S$, a partial ranking $\succ_j$ over $T_j \subseteq Q$; 
Target student $s_i \in S$; 
Set of questions $L_i$ to rank for $s_i$.\\
\textbf{OUTPUT:} a partial order $\hat{\succ_i}$ over $L_i$.
\begin{algorithmic}[1]
\Function{EduRank}{$S,Q,\succ,i,L_i$}
   \ForAll {$q\in L_i$}
	\State $c(q) =  \sum_{q_l\in L_i \setminus q} rv(q, q_l,S)$
\EndFor 
\State  
$\hat{\succ_i} \leftarrow \{\forall   (q_k,q_l) \in \binom{L_i}{2},
q_k \hat{\succ_i}  q_l \text{ iff }
  c(q_k) > c(q_l)\}$
\State \Return $\hat{\succ_i}$ 
\EndFunction
\end{algorithmic}
\caption{Psudeocode for the EduRank algorithm}
  \label{alg:edurank}%_
\end{figure}

\section{Empirical Evaluation}
\label{sec:empirical}

We now  describe a set of experiments comparing EduRank to other algorithms on the difficulty ranking problem. We describe the datasets that were used and our method for defining a difficulty ranking, then we discuss the performance of the various algorithms.

\subsection{Datasets}

We conducted experiments on two real world educational datasets. The first dataset was published in the KDD cup 2010 by the Pittsburgh Science of Learning Center (PSLC)~\footnote{\url{https://pslcdatashop.web.cmu.edu/KDDCup}} \cite{koedinger2010data}. We used the Algebra 1 dataset from the competition, containing about 800,000 answering attempts by 575 students, collected during 2005-2006.  Data sparsity in this dataset (the amount of empty cells in the student/question matrix) was $99.29\%$. We used the following features for each question: question ID, 
the number of retries needed to solve the problem by the student,
 and the duration of time required by the student to submit the answer. Other features in this dataset were not used
 in the study. If the number of retries needed to solve the problem was 0, this means the students solved the problem on a first attempt (we refer to this event as a {\em correct first attempt}).

The second dataset, which we call K12, is an unpublished dataset obtained from an e-learning system installed in 120 schools and used by more than 10,000 students. The records in this dataset were anonymized and approved by the institutional review board of the Ben-Gurion university. This dataset contains about 900,000 answering attempts in various topics including mathematics, English as a second language, and social studies.  The data sparsity in this dataset was $98.24\%$. We used the following features for each question: question ID, the answer provided by the student and the associated grade for each attempt to solve the question.  Unfortunately, this dataset does not contain time stamps for each provided response, so we cannot compute the duration of time until a question was answered.

\subsection{Computing the Difficulty Ranking}

EduRank assumes that each student has a personal difficulty ranking over questions.  In this section we show how we inferred this ranking from the features in the dataset. 
An obvious candidate for the difficulty ranking are the grades that the student got on each question. There are several reasons however as to why grades are an insufficient measurement of difficulty. First, in most questions in the PSLC dataset, the  final grade is either 0 or 1. There were a number of multiple choice questions (between 3 and 4 possible answers) in the datasets, but the dichotomy between low and high grades was also displayed here. To understand this dichotomy, note that students were allowed to repeat the question until they succeeded. It is not surprising that after several retries most students were able to identify the correct answer. A zero grade for a question occurs most often when it was not attempted by the student more than once. 

%One  approach is to consider additional features in addition to grades (or correct first attempts), that are present in the datasets, and which correlate with the difficulty of the question for the individual student.  Specifically, 
%Other 

Two alternative  families of methods for estimating item difficulty
are  {\em item response theory} (IRT)~\cite{wauters2012item} and {\em Bayesian knowledge  tracing} (BKT)~\cite{pardos2011kt}. The IRT model assumes that many students have completed a
test of dichotomous items and assigns each student a proficiency parameter.  It models variation of student proficiency 
across different items. Bayesian Knowledge Tracing captures dynamic changes in student capabilities over time. Both of these 
models do not take into account the number of retries and response time or reason about similarity between students. 

We assumed that questions that were answered correctly on a first attempt were easier for the student, while questions that required multiple attempts were harder. We also assumed that questions that required more solution time, as registered in the log, were more difficult to the students.   These two properties are not perfect indicators of question difficulty for the student. Indeed, it may occur in multiple choice questions that a student guessed the correct answer on the first attempt, even though the question was quite difficult. We also do not account for {\em gaming the system} strategies that have been modeled in past Interactive Learning Environments work~\cite{baker2008students}. It may also be the case that the length of time reported by the system represents idle time for the student who was not even interacting with the e-learning software, or simply taking a break.   
 However, as we demonstrate later in this section, using grades, number of attempts and response times provide a reasonable approach towards ranking questions.

We   describe the following method for identifying the difficulty ranking. We begin by ranking questions by grades. In the PSLC dataset we use ``correct first attempt'' for this, and in the K12 dataset we   use the grade that the student got on her first attempt. After ranking by grade, we break ties by using the number of attempts that the student took  before submitting  the correct answer. When the student did not achieve a correct answer we use all the attempts that the student has made. Then, we break ties again on the PSLC dataset using the elapsed time.

\begin{figure}
\centering
\subfigure[Grades]{
   \includegraphics[width=8.5cm]{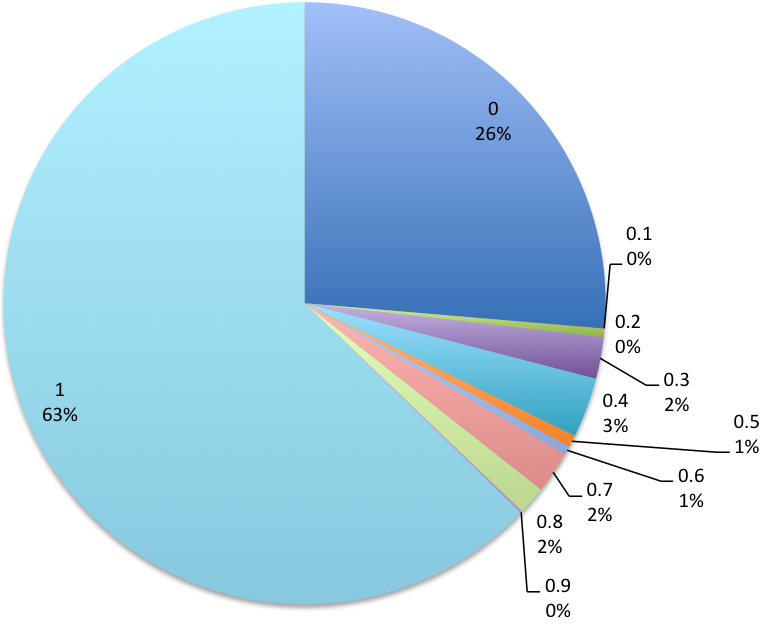}
}
\subfigure[Difficulty Ranking]{  \includegraphics[width=7.5cm]{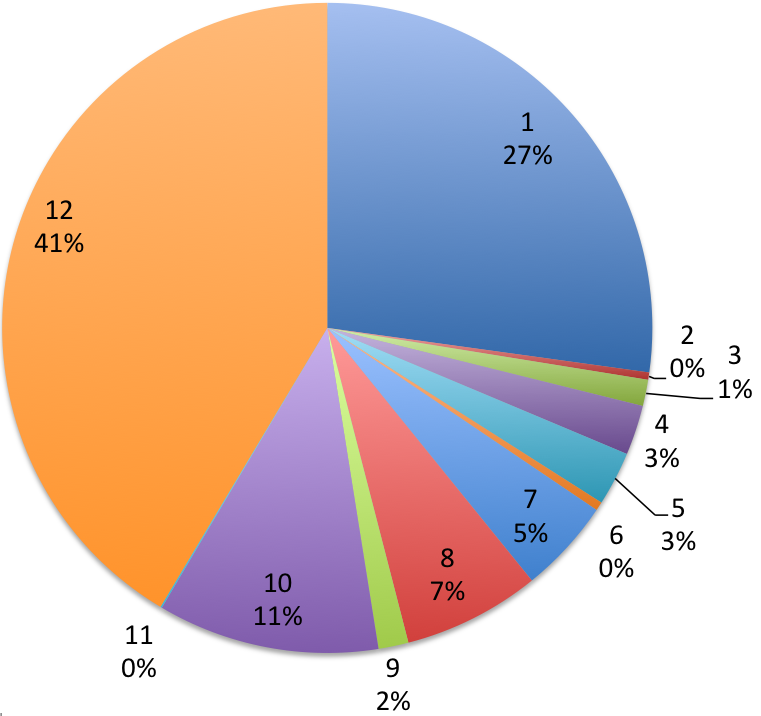}
}
\caption{Distribution over grades and difficulty ranking positions for K12 dataset}
\label{fig:diffHistogram}
\end{figure}

To demonstrate that, in general, these properties provide a reasonable estimation for the difficulty of the question, Figure~\ref{fig:diffHistogram} shows a distribution over students' grades (top) and positions in the inferred difficulty ranking which considered grades and retries (bottom).  Note that the different values for grades represent answers to multiple select questions. For example, a grade of 0.9 will be achieved when 9/10 correct answers were selected by the student.  As can be clearly seen from the figure, there are substantially more classes in the difficulty ranking when adding additional features.

To motivate the EduRank approach, we show that students exhibited a wide degree of variance over difficulty rankings when solving questions.   We chose four
  math topics at random from the dataset, and compared the rankings of
  all students that solved all questions for each topic using the AP
  metric defined in Equation~\ref{eq:4}. In each topic, the AP metric
  is calculated by choosing one student at random and comparing all
  other students' rankings to this student ranking.
  Table~\ref{tab:onesize} presents the topics tested in the K12
  datasets. For each such topic we give the topic ID, the number of
  questions belonging to this topic and the calculated AP metric. As
  can be seen from the rightmost column in the table, all AP values
  are significantly lower than 1, which is the   value for
  perfect similarity between   difficulty rankings. 
  This means that  difficulty rankings with other students are not well correlated. It  supports  the approach to sequence  
  questions to students in a personal manner as opposed to a ``one size fits all" approach.

\begin{table}[t]
  \centering
 \begin{tabular}{|c|c|c|}
\hline
Topic & Num. & AP\\
& questions & \\
\hline
fractions & 33 & 0.31 \\
powers & 48 & 0.36\\
rectangles & 17 & 0.51\\
circles & 61 & 0.34\\
  \hline
 \end{tabular}
 \caption{AP metric  measuring students' difficulty rankings for questions in different topics}
   \label{tab:onesize}
\end{table}

\subsection{Methods}

We used the two ranking scoring metrics  that we described earlier --- NDPM and AP. Many papers in information retrieval also report NDCG, which is a ranking metric for datasets where each item has a score, and thus measures the difference in scores when ranking errors occur. In our case, where we do not have meaningful scores, only pure rankings, NDCG is less appropriate~\cite{kanoulas2009empirical}.  

We compared the performance of a number of alternative  methods to EduRank. First, we compared to the original EigenRank algorithm, which differs from EduRank in that the  similarity metric between users and aggregation method  is only based on grades.  In the K12 dataset we also compared to the default ranking method  provided  already used  in  the system. This method (denoted CER) ranks questions according to increasing order of difficulty as determined by the domain experts.  Second, we  used two popular collaborative filtering methods  that rank by decreasing predicted scores --- a memory-based user-user KNN method using the Pearson correlation (denoted UBCF for user based collaborative filtering), and a {\em matrix factorization} (MF) method using SVD (denoted SVD) to compute latent factors of items and users~\cite{breese1998empirical,zhou2008large,schatten2015integration}. In both cases we used the Mahout\footnote{\url{https://mahout.apache.org/}} implementation of the algorithms \cite{schelter2012collaborative}.  
 Finally, we implemented an approach that assigns questions based on a students' average score over topics. This method follows works  that consider the student's mastery level of a topic when predicting performance~\cite{corbett1994knowledge}. We computed the average score  that the student got for all  questions that  belong to the same  topic. We then rank the topics by decreasing average score, and rank the questions by the topic they belong to. We denote this method the {\em topic-based ranker} (TBR). This measure was used only on the K12 dataset where we have available topic data.

The collaborative filtering algorithms described above all require an item score as an input. We computed scores as follows: We began with the grade (first attempt) that the user got on a question, normalized to the $[0-1]$ range. For each retry of the question we reduced this grade by 0.2 points in accordance with guidelines from the K12 educational expert team. For the PSLC dataset, 
we reduced the (normalized) elapsed time solving the question (of each attempt) from the score. The elapsed time is normalized to the scale $[0-1]$. In both cases, any negative score is converted to zero. 

\begin{figure}
\centering
\subfigure[AP score (higher is better)]{
  \includegraphics[width=8.6cm]{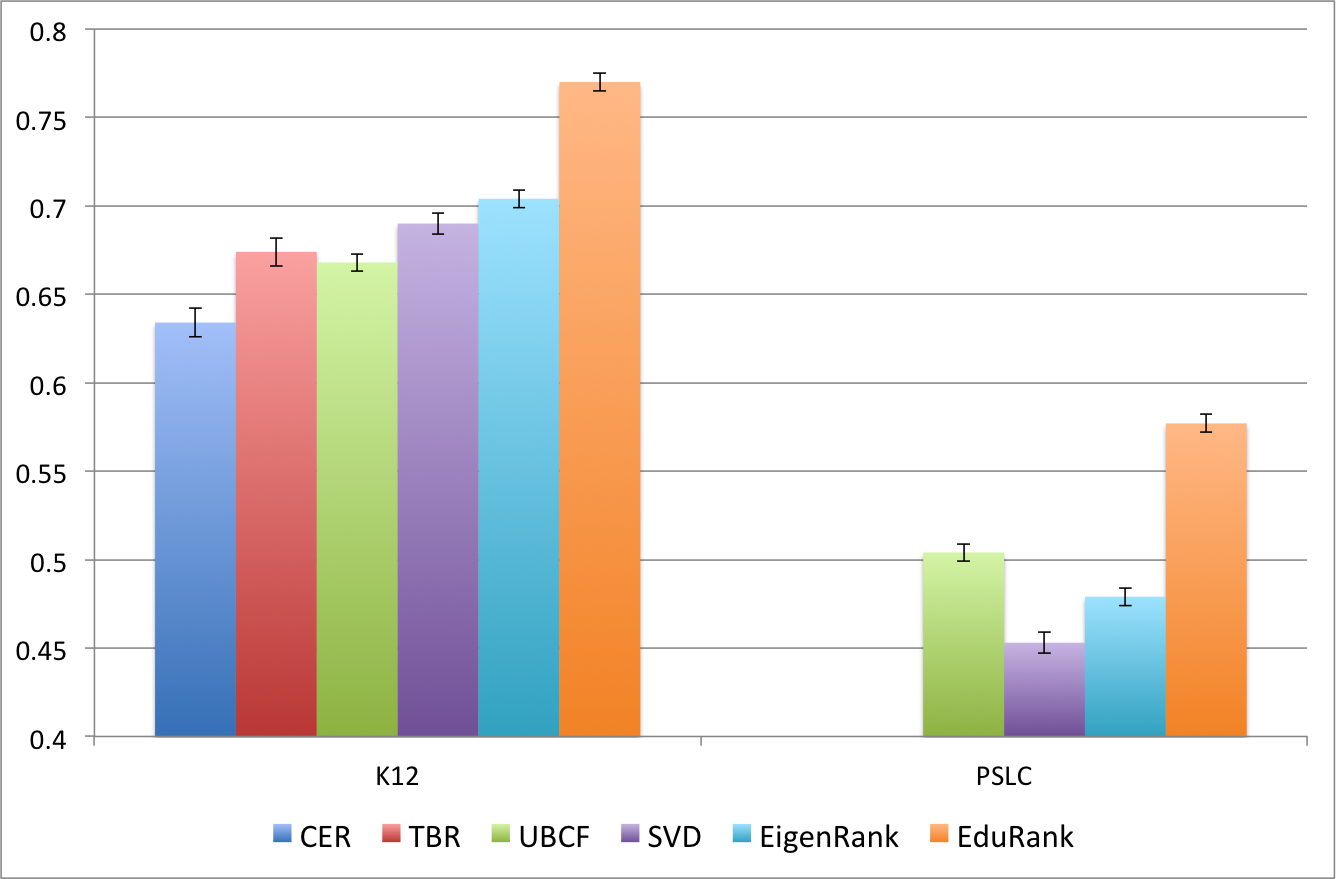}
}
%\qquad
\subfigure[NDPM score (lower is better)]{
  \includegraphics[width=8.3cm]{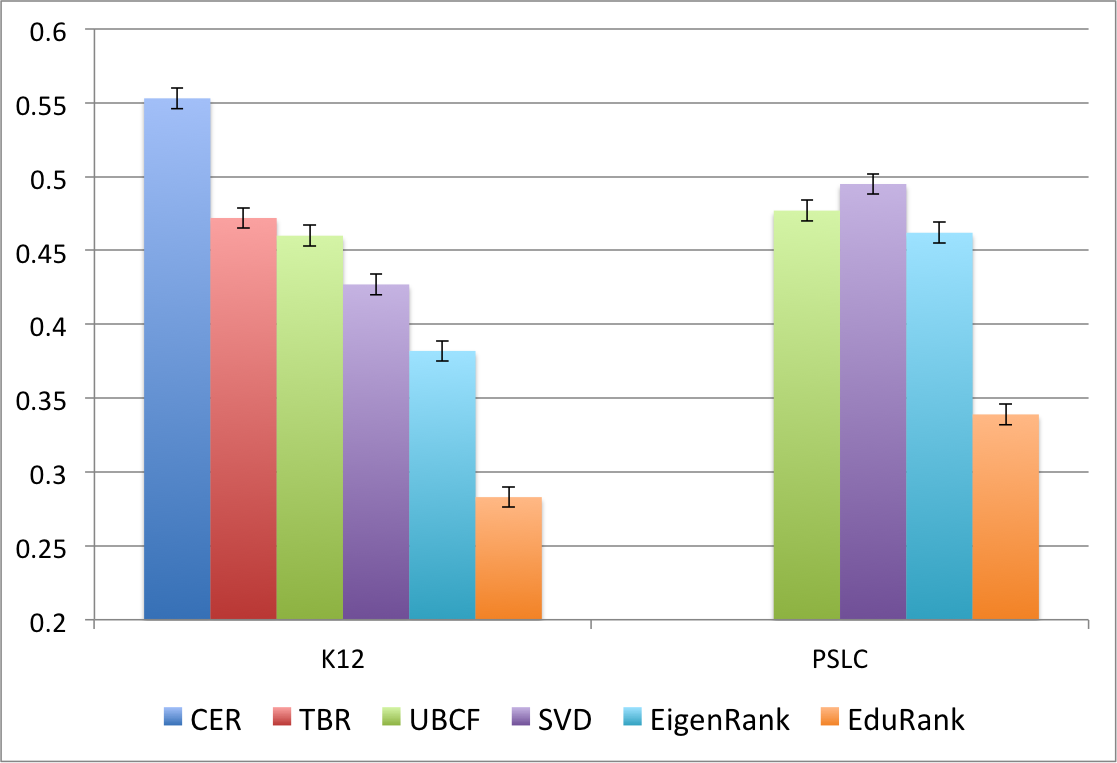}
}
  \caption{Performance Comparison (error bars represent the 95\% confidence interval)}
  \label{fig:fscoreap}
\end{figure}

\subsection{Results}
We ran the following experiment--- for each student $s_i$ we split her
answered questions into two sets of equal size: a training set $T_i$,
which is given as input to the various algorithms, and a test set
$L_i$ that the algorithms must rank. The split is performed according
to the time stamp of the answers. Earier answers are in the training set, while later answers are in the test
set. We then compare the result of each algorithm to the difficulty
ranking explained above using NDPM and AP. The AP metric is also used
to measure similarity between neighboring students in EduRank.  We
note that (1) it is standard practice in ML to use the same metric in
the algorithm and the evaluation, and (2) the AP measure was computed
over the training set in the algorithm, but over the test set in the
evaluation. Notice that for NDPM, the lower the score, the better the
ranking, while for AP, better rankings result in higher scores.  For
all approaches, we ordered the questions in $L_i$ by decreasing order of
difficulty (harder questions were ranked higher in the list).

As can be seen in Figure~\ref{fig:fscoreap}, EduRank is better than
all other approaches on both datasets using both metrics. The results
are statistically significant ($p < 0.05$, paired t-test between
EduRank and the leading competing algorithm).

Looking at the other collaborative filtering methods we can see that
EigenRank and UBCF present comparable performance. This is not very
surprising, because these 2 methods do not take as input a ranking,
but an item score, as we explain above. As the score is only a proxy
to the actual ranking, it is no surprise that these algorithms do not
do as well in predicting the true difficulty ranking.

% \commentout{
% We also compared the algorithm to CofiRank~\cite{weimer2009maximum},
% which uses maximum margin matrix factorisation to optimize rankings of
% items for collaborative filtering. This algorithm takes ranking as
% input. On the K12 data set, EdurRank achieved an AP measure of 0.80,
% compared to 0.77 of CofiRank ($p<0.002$). On the K9 dataset, EduRank
% achieved an AP measure of 0.632, compared to 0.626 of CofiRank. For
% this dataset, there was no We used a freely available API for running
% CofiRank which did not provide AP scores.
% }
Of the non-CF methods, TBR does relatively well. Our intuition is that
identifying the student mastery level in topics is an important factor
in establishing the difficulty of a question for that particular
student. It is hence interesting to investigate in future research how
EduRank can also benefit from the information encapsulated in
topics. Nonetheless TBR can be too limiting in practice, because when
a teacher wants to create a practice assignment in a particular topic,
perhaps one that the student has not yet mastered, then TBR cannot be
used to rank questions within that topic.

The method that performed the worst is the content expert ranking
(CER). This is especially interesting as this is the only information
that is currently available to teachers using the K12 e-learning
system for deciding on the difficulty of questions. There can be two
sources to this sub-optimal performance. First, it may be that it is
too hard, even for experts, to estimate the difficulty of a question
for students. Second, this may be an evidence that personalizing the
order of questions for a particular student is important for
this application.

Producing personalized difficulty rankings for students may be needed many times when used in a classroom, both for multiple students, and for different sessions of the same student. As such, it is critical to choose algorithms that can operate within reasonable time constraints. Table~\ref{table:runtime} shows the execution time of each algorithm for building the models and computing the recommended rankings. The dataset used is the K12 dataset with 918,792 answered questions. We report the overall time for computing a single difficulty ranking for all students and per student.
Our experiments were conducted on a Mac Book Air 1.7GHz Intel Core i7 with 8GB RAM.

CER is obviously the fastest algorithm, as it requires no computation, only the retrieval of the fixed question difficulty level from the database. The memory-based algorithms are next, with
UBCF being the fastest among the remaining algorithms, followed by EduRank and EigenRank. The SVD algorithm, requiring the construction of the matrix factorization model is the slowest here. That being said, as the number of students and questions grow, model-based approaches are expected to work faster than memory-based. 
In that case, we can move some computations to an initialization phase. For example, we can compute the similarity between students offline, and cache the top $k$ nearest neighbors, reducing the online computation time only to the aggregation of the rankings of similar users.

%\if 0
% Table generated by Excel2LaTeX from sheet 'Copact Skills'
\begin{table}[t]
  \centering
    \scalebox{0.31}
{
    \begin{tabular}{rcrrrrrrrr}
%    \toprule
    \multicolumn{2}{c}{\textbf{Gold Standard}} & \multicolumn{2}{c}{\textbf{EduRank Ranking}} & \multicolumn{2}{c}{\textbf{EigenRank Ranking}} & \multicolumn{2}{c}{\textbf{UBCF Ranking}} & \multicolumn{2}{c}{\textbf{SVD Ranking}} \\
%    \midrule
    \multicolumn{1}{c}{\textbf{}} & \textbf{} &       &       &       &       &       &       &       &  \\
    \multicolumn{1}{c}{\textbf{}} & \textbf{} &       &       &       &       &       &       &       &  \\
    \multicolumn{1}{c}{KC} & True Rank & \multicolumn{1}{c}{KC} & \multicolumn{1}{c}{True Rank} & \multicolumn{1}{c}{KC} & \multicolumn{1}{c}{True Rank} & \multicolumn{1}{c}{KC} & \multicolumn{1}{c}{True Rank} & \multicolumn{1}{c}{KC} & \multicolumn{1}{c}{True Rank} \\
    \multicolumn{1}{l}{\textbf{Order of Operations, choose options }} & \textbf{1} & \multicolumn{1}{l}{\textbf{Order of Operations, choose options }} & \multicolumn{1}{c}{\textbf{1}} & \multicolumn{1}{l}{\textbf{Order of Operations, Brackets}} & \multicolumn{1}{c}{\textbf{7}} & \multicolumn{1}{l}{\textbf{Multiply, Equals 54}} & \multicolumn{1}{c}{\textbf{12}} & \multicolumn{1}{l}{\textbf{Multiply, Big Numbers}} & \multicolumn{1}{c}{\textbf{4}} \\
    \multicolumn{1}{l}{\textbf{Letters Order}} & \textbf{1} & \multicolumn{1}{l}{\textbf{Natural Numbers, Verbal Claims}} & \multicolumn{1}{c}{\textbf{3}} & \multicolumn{1}{l}{\textbf{Natural numbers, In between}} & \multicolumn{1}{c}{\textbf{12}} & \multicolumn{1}{l}{\textbf{Multiply, Choose Between 2 }} & \multicolumn{1}{c}{\textbf{12}} & \multicolumn{1}{l}{\textbf{Multiply, Bigger than}} & \multicolumn{1}{c}{\textbf{10}} \\
    \multicolumn{1}{l}{\textbf{Multiply, Equals 40 }} & \textbf{2} & \multicolumn{1}{l}{\textbf{Add, Sub, Equals 30}} & \multicolumn{1}{c}{\textbf{10}} & \multicolumn{1}{l}{\textbf{Div, No Mod, Mod 1}} & \multicolumn{1}{c}{\textbf{11}} & \multicolumn{1}{l}{\textbf{Multiply, Bigger than}} & \multicolumn{1}{c}{\textbf{10}} & \multicolumn{1}{l}{\textbf{Order of Operations, Brackets}} & \multicolumn{1}{c}{\textbf{5}} \\
    \multicolumn{1}{l}{\textbf{Natural Numbers, Verbal Claims}} & \textbf{3} & \multicolumn{1}{l}{\textbf{Letters Order}} & \multicolumn{1}{c}{\textbf{1}} & \multicolumn{1}{l}{\textbf{Div, Div and Mod}} & \multicolumn{1}{c}{\textbf{11}} & \multicolumn{1}{l}{\textbf{Div, No Mod, Mod 1}} & \multicolumn{1}{c}{\textbf{11}} & \multicolumn{1}{l}{\textbf{Order of Operations, Equals 5}} & \multicolumn{1}{c}{\textbf{6}} \\
    \multicolumn{1}{l}{\textbf{Multiply, Big Numbers}} & \textbf{4} & \multicolumn{1}{l}{\textbf{Add, Sub, Verbal Claims}} & \multicolumn{1}{c}{\textbf{7}} & \multicolumn{1}{l}{\textbf{Multiply, Big Numbers}} & \multicolumn{1}{c}{\textbf{7}} & \multicolumn{1}{l}{\textbf{Div, No Mod, Mod 2}} & \multicolumn{1}{c}{\textbf{12}} & \multicolumn{1}{l}{\textbf{Natural Numbers, Verbal Claims}} & \multicolumn{1}{c}{\textbf{3}} \\
    \multicolumn{1}{l}{\textbf{Order of Operations, Brackets}} & \textbf{5} & \multicolumn{1}{l}{\textbf{Order of Operations, Equals 5}} & \multicolumn{1}{c}{\textbf{6}} & \multicolumn{1}{l}{\textbf{Div, Exists?}} & \multicolumn{1}{c}{\textbf{8}} & \multicolumn{1}{l}{\textbf{Multiply, Big Numbers}} & \multicolumn{1}{c}{\textbf{4}} & \multicolumn{1}{l}{\textbf{Add, Sub, Equals 30}} & \multicolumn{1}{c}{\textbf{10}} \\
    \multicolumn{1}{l}{\textbf{Zero, Equals Zero}} & \textbf{5} & \multicolumn{1}{l}{\textbf{Order of Operations, Brackets}} & \multicolumn{1}{c}{\textbf{5}} & \multicolumn{1}{l}{\textbf{Multiply, Equals 40 }} & \multicolumn{1}{c}{\textbf{2}} & \multicolumn{1}{l}{\textbf{Natural Numbers, Verbal Claims}} & \multicolumn{1}{c}{\textbf{3}} & \multicolumn{1}{l}{\textbf{Order of Operations, Brackets}} & \multicolumn{1}{c}{\textbf{7}} \\
    \multicolumn{1}{l}{\textbf{Order of Operations, Equals 5}} & \textbf{6} & \multicolumn{1}{l}{\textbf{Zero, Equals Zero}} & \multicolumn{1}{c}{\textbf{5}} & \multicolumn{1}{l}{\textbf{Div, Mod 2}} & \multicolumn{1}{c}{\textbf{12}} & \multicolumn{1}{l}{\textbf{Order of Operations, choose options }} & \multicolumn{1}{c}{\textbf{1}} & \multicolumn{1}{l}{\textbf{Div, Mod 2}} & \multicolumn{1}{c}{\textbf{12}} \\
    \multicolumn{1}{l}{\textbf{Order of Operations, Brackets}} & \textbf{7} & \multicolumn{1}{l}{\textbf{Multiply, Big Numbers}} & \multicolumn{1}{c}{\textbf{4}} & \multicolumn{1}{l}{\textbf{Multiply, Choose between 2}} & \multicolumn{1}{c}{\textbf{12}} & \multicolumn{1}{l}{\textbf{Order of Operations, Equals 5}} & \multicolumn{1}{c}{\textbf{6}} & \multicolumn{1}{l}{\textbf{Add, Sub, Verbal Claims}} & \multicolumn{1}{c}{\textbf{7}} \\
    \multicolumn{1}{l}{\textbf{Add, Sub, Verbal Claims}} & \textbf{7} & \multicolumn{1}{l}{\textbf{Div, Mod 2}} & \multicolumn{1}{c}{\textbf{12}} & \multicolumn{1}{l}{\textbf{Order of Operations, Which is bigger}} & \multicolumn{1}{c}{\textbf{11}} & \multicolumn{1}{l}{\textbf{Multiply, Choose between 2}} & \multicolumn{1}{c}{\textbf{12}} & \multicolumn{1}{l}{\textbf{Order of Operations, choose options }} & \multicolumn{1}{c}{\textbf{1}} \\
    \multicolumn{1}{l}{\textbf{Multiply, Big Numbers}} & \textbf{7} & \multicolumn{1}{l}{\textbf{Div, No Mod, Mod 2}} & \multicolumn{1}{c}{\textbf{12}} & \multicolumn{1}{l}{\textbf{Order of Operations, Brackets}} & \multicolumn{1}{c}{\textbf{5}} & \multicolumn{1}{l}{\textbf{Multiply, Choose between 2}} & \multicolumn{1}{c}{\textbf{12}} & \multicolumn{1}{l}{\textbf{Multiply, Equals 54}} & \multicolumn{1}{c}{\textbf{12}} \\
    \multicolumn{1}{l}{\textbf{Div, Exists?}} & \textbf{8} & \multicolumn{1}{l}{\textbf{Order of Operations, Brackets}} & \multicolumn{1}{c}{\textbf{7}} & \multicolumn{1}{l}{\textbf{Div, Mod 1}} & \multicolumn{1}{c}{\textbf{11}} & \multicolumn{1}{l}{\textbf{Order of Operations, Brackets}} & \multicolumn{1}{c}{\textbf{7}} & \multicolumn{1}{l}{\textbf{Div, Exists?}} & \multicolumn{1}{c}{\textbf{12}} \\
    \multicolumn{1}{l}{\textbf{Substruction}} & \textbf{9} & \multicolumn{1}{l}{\textbf{Order of Operations, Which is bigger}} & \multicolumn{1}{c}{\textbf{11}} & \multicolumn{1}{l}{\textbf{Order of Operations, only \%, /}} & \multicolumn{1}{c}{\textbf{11}} & \multicolumn{1}{l}{\textbf{Order of Operations, Brackets}} & \multicolumn{1}{c}{\textbf{5}} & \multicolumn{1}{l}{\textbf{Div, No Mod, Mod 2}} & \multicolumn{1}{c}{\textbf{12}} \\
    \multicolumn{1}{l}{\textbf{Multiply, Bigger than}} & \textbf{10} & \multicolumn{1}{l}{\textbf{Order of Operations, only \%, /}} & \multicolumn{1}{c}{\textbf{11}} & \multicolumn{1}{l}{\textbf{Polygon, Parallel sides}} & \multicolumn{1}{c}{\textbf{10}} & \multicolumn{1}{l}{\textbf{Letters Order}} & \multicolumn{1}{c}{\textbf{1}} & \multicolumn{1}{l}{\textbf{Multiply, Big Numbers}} & \multicolumn{1}{c}{\textbf{7}} \\
    \multicolumn{1}{l}{\textbf{Add, Sub, Equals 30}} & \textbf{10} & \multicolumn{1}{l}{\textbf{Multiply, Big Numbers}} & \multicolumn{1}{c}{\textbf{7}} & \multicolumn{1}{l}{\textbf{Letters Order}} & \multicolumn{1}{c}{\textbf{1}} & \multicolumn{1}{l}{\textbf{Rectangle, Identify}} & \multicolumn{1}{c}{\textbf{12}} & \multicolumn{1}{l}{\textbf{Natural numbers, In between}} & \multicolumn{1}{c}{\textbf{12}} \\
    \multicolumn{1}{l}{Polygon, Parallel sides} & 10    & \multicolumn{1}{l}{Div, Exists?} & \multicolumn{1}{c}{12} & \multicolumn{1}{l}{Order of Operations, Equals 5} & \multicolumn{1}{c}{6} & \multicolumn{1}{l}{Multiply, Big Numbers} & \multicolumn{1}{c}{7} & \multicolumn{1}{l}{Zero, Equals Zero} & \multicolumn{1}{c}{5} \\
    \multicolumn{1}{l}{Order of Operations, only +, - } & 11    & \multicolumn{1}{l}{Substruction} & \multicolumn{1}{c}{9} & \multicolumn{1}{l}{Substruction} & \multicolumn{1}{c}{9} & \multicolumn{1}{l}{Polygon, Identify} & \multicolumn{1}{c}{12} & \multicolumn{1}{l}{Order of Operations, Which is bigger} & \multicolumn{1}{c}{11} \\
    \multicolumn{1}{l}{Order of Operations, only \%, /} & 11    & \multicolumn{1}{l}{Polygon, Parallel sides} & \multicolumn{1}{c}{10} & \multicolumn{1}{l}{Add, Sub, Verbal Claims} & \multicolumn{1}{c}{7} & \multicolumn{1}{l}{Zero, Equals Zero} & \multicolumn{1}{c}{5} & \multicolumn{1}{l}{Div, Div and Mod} & \multicolumn{1}{c}{11} \\
    \multicolumn{1}{l}{Order of Operations, Which is bigger} & 11    & \multicolumn{1}{l}{Order of Operations, only +, - } & \multicolumn{1}{c}{11} & \multicolumn{1}{l}{Multiply, Big Numbers} & \multicolumn{1}{c}{4} & \multicolumn{1}{l}{Order of Operations, only +, - } & \multicolumn{1}{c}{11} & \multicolumn{1}{l}{Letters Order} & \multicolumn{1}{c}{1} \\
    \multicolumn{1}{l}{Div, Mod 1} & 11    & \multicolumn{1}{l}{Div, No Mod, Mod 1} & \multicolumn{1}{c}{11} & \multicolumn{1}{l}{Natural Numbers, Verbal Claims} & \multicolumn{1}{c}{3} & \multicolumn{1}{l}{Add, Sub, Equals 30} & \multicolumn{1}{c}{10} & \multicolumn{1}{l}{Angles, Find Bigger} & \multicolumn{1}{c}{12} \\
    \multicolumn{1}{l}{Div, Div and Mod} & 11    & \multicolumn{1}{l}{Multiply, Bigger than} & \multicolumn{1}{c}{10} & \multicolumn{1}{l}{Add, Sub, Equals 30} & \multicolumn{1}{c}{10} & \multicolumn{1}{l}{Polygon, Parallel sides} & \multicolumn{1}{c}{10} & \multicolumn{1}{l}{Multiply, Choose between 2} & \multicolumn{1}{c}{12} \\
    \multicolumn{1}{l}{Div, No Mod, Mod 1} & 11    & \multicolumn{1}{l}{Div, Exists?} & \multicolumn{1}{c}{8} & \multicolumn{1}{l}{Order of Operations, choose options } & \multicolumn{1}{c}{1} & \multicolumn{1}{l}{Add, Sub, Verbal Claims} & \multicolumn{1}{c}{7} & \multicolumn{1}{l}{Div, Mod 1} & \multicolumn{1}{c}{11} \\
    \multicolumn{1}{l}{Natural numbers, In between} & 12    & \multicolumn{1}{l}{Div, Mod 1} & \multicolumn{1}{c}{11} & \multicolumn{1}{l}{Order of Operations, only +, - } & \multicolumn{1}{c}{11} & \multicolumn{1}{l}{Div, Mod 1} & \multicolumn{1}{c}{11} & \multicolumn{1}{l}{Multiply, Choose between 2} & \multicolumn{1}{c}{12} \\
    \multicolumn{1}{l}{Multiply, Equals 54} & 12    & \multicolumn{1}{l}{Multiply, Equals 40 } & \multicolumn{1}{c}{2} & \multicolumn{1}{l}{Zero, Equals Zero} & \multicolumn{1}{c}{5} & \multicolumn{1}{l}{Div, Mod 2} & \multicolumn{1}{c}{12} & \multicolumn{1}{l}{Div, No Mod, Mod 1} & \multicolumn{1}{c}{11} \\
    \multicolumn{1}{l}{Multiply, Choose between 2} & 12    & \multicolumn{1}{l}{Div, Div and Mod} & \multicolumn{1}{c}{11} & \multicolumn{1}{l}{Div, No Mod, Mod 2} & \multicolumn{1}{c}{12} & \multicolumn{1}{l}{Div, Div and Mod} & \multicolumn{1}{c}{11} & \multicolumn{1}{l}{Polygon, Parallel sides} & \multicolumn{1}{c}{10} \\
    \multicolumn{1}{l}{Multiply, Choose between 2} & 12    & \multicolumn{1}{l}{Multiply, Choose between 2} & \multicolumn{1}{c}{12} & \multicolumn{1}{l}{Div, Exists?} & \multicolumn{1}{c}{12} & \multicolumn{1}{l}{Order of Operations, only \%, /} & \multicolumn{1}{c}{11} & \multicolumn{1}{l}{Div, Exists?} & \multicolumn{1}{c}{8} \\
    \multicolumn{1}{l}{Div, Mod 2} & 12    & \multicolumn{1}{l}{Multiply, Choose Between 2 } & \multicolumn{1}{c}{12} & \multicolumn{1}{l}{Multiply, Bigger than} & \multicolumn{1}{c}{10} & \multicolumn{1}{l}{Order of Operations, Which is bigger} & \multicolumn{1}{c}{11} & \multicolumn{1}{l}{Order of Operations, only \%, /} & \multicolumn{1}{c}{11} \\
    \multicolumn{1}{l}{Div, Exists?} & 12    & \multicolumn{1}{l}{Rectangle, Identify} & \multicolumn{1}{c}{12} & \multicolumn{1}{l}{Multiply, Choose Between 2 } & \multicolumn{1}{c}{12} & \multicolumn{1}{l}{Div, Exists?} & \multicolumn{1}{c}{8} & \multicolumn{1}{l}{Substruction} & \multicolumn{1}{c}{9} \\
    \multicolumn{1}{l}{Div, No Mod, Mod 2} & 12    & \multicolumn{1}{l}{Polygon, Identify} & \multicolumn{1}{c}{12} & \multicolumn{1}{l}{Rectangle, Identify} & \multicolumn{1}{c}{12} & \multicolumn{1}{l}{Div, Exists?} & \multicolumn{1}{c}{12} & \multicolumn{1}{l}{Order of Operations, only +, - } & \multicolumn{1}{c}{11} \\
    \multicolumn{1}{l}{Angles, Find Bigger} & 12    & \multicolumn{1}{l}{Multiply, Equals 54} & \multicolumn{1}{c}{12} & \multicolumn{1}{l}{Multiply, Equals 54} & \multicolumn{1}{c}{12} & \multicolumn{1}{l}{Natural numbers, In between} & \multicolumn{1}{c}{12} & \multicolumn{1}{l}{Multiply, Equals 40 } & \multicolumn{1}{c}{2} \\
    \multicolumn{1}{l}{Angles, Find Bigger} & 12    & \multicolumn{1}{l}{Angles, Find Bigger} & \multicolumn{1}{c}{12} & \multicolumn{1}{l}{Angles, Find Bigger} & \multicolumn{1}{c}{12} & \multicolumn{1}{l}{Substruction} & \multicolumn{1}{c}{9} & \multicolumn{1}{l}{Angles, Find Bigger} & \multicolumn{1}{c}{12} \\
    \multicolumn{1}{l}{Rectangle, Identify} & 12    & \multicolumn{1}{l}{Angles, Find Bigger} & \multicolumn{1}{c}{12} & \multicolumn{1}{l}{Angles, Find Bigger} & \multicolumn{1}{c}{12} & \multicolumn{1}{l}{Multiply, Equals 40 } & \multicolumn{1}{c}{2} & \multicolumn{1}{l}{Multiply, Choose Between 2 } & \multicolumn{1}{c}{12} \\
    \multicolumn{1}{l}{Polygon, Identify} & 12    & \multicolumn{1}{l}{Natural numbers, In between} & \multicolumn{1}{c}{12} & \multicolumn{1}{l}{Multiply, Choose between 2} & \multicolumn{1}{c}{12} & \multicolumn{1}{l}{Angles, Find Bigger} & \multicolumn{1}{c}{12} & \multicolumn{1}{l}{Polygon, Identify} & \multicolumn{1}{c}{12} \\
    \multicolumn{1}{l}{Multiply, Choose Between 2 } & 12    & \multicolumn{1}{l}{Multiply, Choose between 2} & \multicolumn{1}{c}{12} & \multicolumn{1}{l}{Polygon, Identify} & \multicolumn{1}{c}{12} & \multicolumn{1}{l}{Angles, Find Bigger} & \multicolumn{1}{c}{12} & \multicolumn{1}{l}{Rectangle, Identify} & \multicolumn{1}{c}{12} \\
%    \bottomrule
    \end{tabular}%
 }
  \caption{Rankings outputted by the different algorithms for a sample target student}
\label{table:results36}
\end{table}%
%\fi
% Table generated by Excel2LaTeX from sheet 'Run Time'
%\if 0
\begin{table}
  \centering
    \begin{tabular}{|l|c|c|}
%    \toprule
\hline
    Algorithm  & Run Time (Sec) & Time per Student (millisec) \\
    \hline
 %   \midrule
   CER & 197.6  & 19.2\\
   UBCF & 445.2 & 43.2 \\
    TBR & 625.2 & 60.6\\
   EduRank & 631.8 & 61.2\\
    EigenRank & 795.9 & 77.2 \\
   SVD  & 1490 & 144.4\\
    \hline
  %  \bottomrule
    \end{tabular}%
  \caption{Execution Time}
  \label{table:runtime}%
\normalsize
\end{table}%

%\fi

\subsection{Case Study}

To   demonstrate the behaviour of the various algorithms, we present the results of the algorithms for a particular student from the K12 dataset. Table~\ref{table:results36}  presents a list of 34 test questions for this student  and the rankings 
that were outputted by the different algorithms, in decreasing order of difficulty. The 15 most difficult questions appear in bold. Each question is denoted by (1) its knowledge component (KC) which was determined by a domain expert (this information was not in the database and the algorithms did not use it), and (2)  the position of the question in the true difficulty ranking - the gold standard - of the student (as computed by the grade of the student and her number of retries when solving the question). This gold standard was used by the NDPM and AP metrics as a reference ranking to judge the performance of all algorithms.  As shown in the table, question types involving ``multiplication of big numbers'' and ``order of operations''  appear prominently in the 15-most difficulty list,  while questions in topics of geometry (``rectangles'', ``polygons'') were easier for the student.

The other columns in the table show the suggested rankings by the various algorithms. For each algorithm, we present the ranking location of each question, and the true ranking of this question as obtained from the gold standard. As can be seen from the results, for this particular student, the UBCF algorithm performed poorly, placing many easy questions for the student at high positions in the ranking (e.g., ``Multiply Eq 54'' which appears at the top of the list but is ranked 12th in the gold standard, and ``div mod'' appears in 4th position in the list and ranked 11th in the gold standard).
 The EigenRank and SVD algorithms demonstrated better results, but still failed to place the most difficult question for the student (e.g., order of operations) at the top of the ranked list. Only the EduRank algorithm was able to place the questions with ``multiplication of big numbers'' and ``order of operation''  type problems in the top 15 list, providing the best personalized difficulty ranking for this student.

\subsection{Addressing the Cold Start Problem}

In a real classroom  new  students join the system on an ongoing basis. At the onset, the system has little to no information on these students.  This is known as the {\em cold start} problem in the  recommendation systems literature~\cite{Schein:2002:MMC:564376.564421}. 
To be able to use EduRank in a real classroom, we need to overcome situations in which  there are not enough questions completed by a target student  to search for similarities in the data set. 

% it may not be possible to  find similar students and compute the ranking prediction for the target student. 
% cold start problem arises when

To tackle this problem, we incorporated  a prior score  $pr(q_k)$ for every question $q_k$ in the training set  by averaging over the scores  of all students that solved this question in the training set. 
The difference between the prior score for a question pair $(q_k,q_l)$  can be used as a proxy for the pair-wise ranking between these two questions. We use a linear combination of the prior score and the student similarity score. To this end we replace Equation~\ref{eq:a9} used by the EduRank algorithm to compute the relative voting $rv(q_k, q_l,S)$ between the  two questions $q_k$ and $q_l$ in the test set $S$ 
given a training set $T_i$ of questions for the target student $i$ by the following equation:
\begin{equation}
  \label{eq:p1}
  \begin{aligned}
 rv(q_k, q_l,S) =&  \alpha_{k,l} *sign(\sum_{j \in S\setminus i} s_{AP}(T_i,\succ_i,\succ_j) \cdot  \gamma(q_k,
  q_l,\succ_j)) + \\ 
  &(1.0 - \alpha_{k,l})*sign(pr(q_k) - pr(q_l))
  \end{aligned}
\end{equation}
{The value of $\alpha_{k,l}$ is set to the number of available rankings of $(q_k, q_l)$  in the training set divided by the neighboring size used by the algorithm when selecting similar students. That is, as the number of students that have solved both $q_k$ and $q_l$ increases, the weight of the prior is reduced.
Hence, as the amount of ranking information that is available in the training set increases, we rely more on this information and less on the computed priors. }
We note that our  solution handles the cold start problem for new students, but not the cold start problem for new questions that are added to the system. 
%We leave discussion of the cold start problem for new questions to future research.

We study the effect of the amount of data collected about students on the prediction performance of  EduRank, and the  extended  algorithm for handling new students, denoted  EduRank+Prior. Table~\ref{table:priorR} demonstrates the training and testing methodology. We varied the  training set to consist of  the first week, first two weeks, first three weeks, first four weeks, first eight weeks and first 40 weeks of records in the system. The test set consists of the 41st week. The table also denotes the percentage of new students in the testing set for the corresponding training set. This  setup was chosen 
to reflect the  the use of the EduRank system in a real classroom, in which  data that is  supplied to the training algorithm from each student increases gradually.

%\if 0
% Table generated by Excel2LaTeX from sheet 'Sheet2'
\begin{table}[t]
  \centering
    \begin{tabular}{|c|l|l|c|}
%    \toprule
  \hline
    \textbf{Experiment} & \textbf{Training Data} & \textbf{Testing Data} & \textbf{\% New Students} \\
  %  \midrule
  \hline
   1& Week 1 & Week 41 & 99\%\\
    2 & Weeks 1-2 & Week 41 & 78\% \\
   3 & Weeks 1-3 & Week 41 & 71\% \\
   4 & Weeks 1-4 & Week 41 & 66\% \\
    5 & Weeks 1-8 & Week 41 & 43\% \\
    6 & Weeks 1-40 & Week 41 & 6\% \\
      \hline
 %   \bottomrule
    \end{tabular}%
  \caption{Training and Testing data for Cold-Start Experiments}
  \label{table:priorR}%
\end{table}%

%\fi 

\begin{figure}[t]
\centering
  \includegraphics[width=8.35cm]{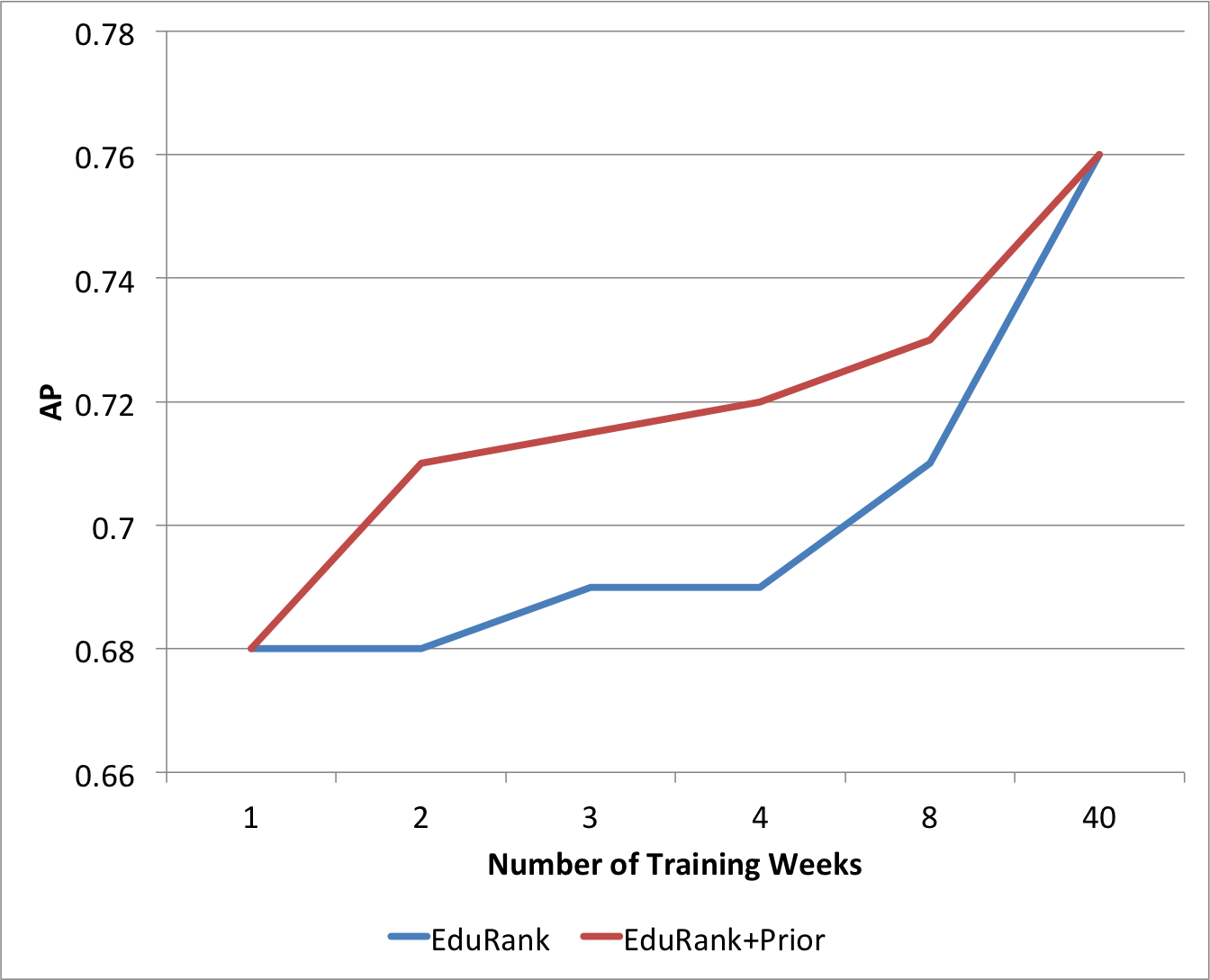}
  \caption{AP for EduRank algorithm  with and without  prior component}
  \label{fig:priorR}
\end{figure}

Figure~\ref{fig:priorR}  shows the results of these experiments, comparing between the algorithm without the prior component and the algorithm with the prior component. We vary the amount of weeks  worth of data in the training set, and  also show the percentage of students in the test set not seen in the training set.
The performance of both models are identical with only 1 week of training data. Then, the EduRank+Prior algorithm  outperforms the basic EduRank algorithm, showing  significant difference as late as the 8th week of training. 
 As more weeks of training data are available, the ratio of new students drop, and EduRank increases the weight of the similarity score over the prior.  When both algorithms gain substantial amount of 40 weeks of record data, their performance is identical as it was in week 1 (no statistically significant difference). 

To show the benefit of using the EduRank+Prior algorithm when dealing with new students, we evaluate the algorithm performance on specific students with low training data. For this test we   randomly chose 50 students and removed most of their training data when predicting their test ranking. Specifically, in each such prediction step, 90\% of training data of each  target student was removed, while the training data of other students was retained and an overall prediction score for the target student was computed. We then 
compared  the performance of  EduRank with and without the prior component on this  data set, 
averaging  the results across all 50 random students.  Our results indicate that the EduRank+Prior algorithm obtained an average AUC score of 0.72 while the EduRank algorithm (without the prior component) obtained an average AUC score of 0.64 for the target students.  

Both of the above results demonstrate that   using a prior-based approach is indeed beneficial for EduRank in cold start scenarios which are to be expected in real world situations.

\section{Deployment in the Classroom}
\label{sec:class}

The previous section evaluated the EduRank approach on offline data. We now report on the  deployment 
 of EduRank in pedagogical context.  The experiment was
conducted during a summer school mathematics session in Israel, from July 21, 2015 to August 15, 2015.\footnote{IRB approval was obtained from Ben-Gurion University.}

The summer school is an elective school for strengthening Math and English as a second language capabilities conducted during the summer holiday. The school is run by professional Math and English teachers. Between classes the students were requested to practice using the K12 system, but this was not mandatory. The teachers and students were unaware of the experiment.

\subsection{Methods}
We compared two methods for sequencing questions to students.
The first approach used the EduRank+Prior algorithm to order
questions for students by increasing order of difficulty, as inferred
by the algorithm.  The second sequencing approach, denoted ASC, was
based on pedagogical experts that created predetermined difficulty
level for each of the questions in the K12 data set ranging from 1
(easiest) to 5 (most difficulty). The sequencing strategy determined
by the pedagogical experts included a set of questions randomly
sampled from the different level of difficulties as follows: 20\%
questions of level 1 difficulty; 30\% questions of level 2 difficulty;
40\% questions of level 3 difficulty; 10\% questions of level 4
difficulty.  The pedagogical experts decided not to include questions
from the most difficult level.  The questions were sequenced to
students in ascending order of difficulty.
Both of the approaches followed  the mastery
learning principle by which knowledge of simple skills should be
demonstrated before moving on to more difficult questions relating to
more complex skills~\cite{block1971mastery} but they differed on how they selected questions for students.

\subsection{Procedure}
We conducted the experiment in two  of the  summer school classes using K12,  one of which was randomly
assigned to use the EduRank+Prior algorithm (denoted EduRank), while the
other classroom was assigned to the ascending (ASC) algorithm.  
Each sequencing algorithm was used in real-time to present questions to students as part of their 
coursework.  After
submitting an answer, students received feedback from the system about
their score for each question. The number of consecutive retries allowed for each
question was limited to three.  Students could also choose not to
answer a question that is given to them by the algorithm.
Students were randomly assigned to the summer school classes and the questions presented by the system matched the topics covered in the classroom. Additionally, the curriculum was identical for the two classes (conditions). The data for the EduRank algorithm was updated each night after class to account for ongoing student learning.
We hypothesized that using the EduRank approach to sequence questions to students would 
lead students  to better  performance on more difficult problems  without reducing their motivation 
in the e-learning  system, as 
compared to the alternative sequencing approach.

Both classes were  administered a pre-test consisting of a single session of 15
questions in mathematics that were sampled from different topics and
expert difficulty levels in mathematics.  The questions for this preliminary
test were chosen by a domain expert. There was no statistically
significant difference between the two groups in each class in the
average score on this preliminary test (ASC condition mean score  was 0.64, STD 16 and EduRank condition mean score was 0.63, STD 18). 
Hence we asserted that students in each group exhibited similar knowledge baselines of the
material.

\subsection{Results}
In the analysis that follows we restricted our attention to sessions of
students who completed the preliminary session.  We did not track
individual students' progress during the study nor control any of the
conditions in the classroom beyond the use of the sequence algorithm.
Students could use the system with no supervision and practice as many
questions as they wanted. Inspecting the two classes in combination, students solved on average 152 questions (stdev 169) and 94\% of students reached and solved questions of levels 4 or 5. During the experiment no student ran out of problems to solve. 

Table~\ref{tab:stats} shows the distribution of the number of students
in each group, the number of questions solved, the average grade (first attempt), and the
average time spent on each question (in seconds).  As shown by the
table, the average grade obtained by students using the EduRank
approach was four points higher than the grade obtained by students
using the sequencing approach. This difference is statistically significance (Repeated Measures ANOVA, $F(1,36361)=18.55,$ \hspace{1mm} $p=1.66\cdot10^{-5}$). When comparing the two
approaches for each level of difficulty (Table~\ref{tab:per}), we can
see that the students using the EduRank approach achieved higher
scores for all questions in each of the levels. Specifically, for easy
questions (level 1) the performance of the EduRank and ASC approach
was identical. However, for more difficult questions (levels 2 and up)
students using the EduRank approach achieved higher performance than
students using the ASC approach. Note that the ASC algorithm did not use questions from level 5. 

Finally, Table~\ref{tab:stats} also shows that students using
the EduRank approach solved more questions and spent more time in the
system than students using the ASC approach. This implies that despite
solving more difficult questions, the students using EduRank were as
 motivated to use the e-learning system as the students using the ASC algorithm.

\begin{table}[t]
    \centering
  \begin{tabular}{|c|c|c|c|c|}
\hline
 Approach  & Num. & Num.  & Avg.  & Time \\
          & Students & Questions & Grade &   (Seconds) \\
\hline
EduRank & 21 & 2027 &  93 & 27.0 \\
ASC & 17 &  1707 &  89 & 17.1\\
\hline
  \end{tabular}
  \caption{Classroom Study Statistics}
  \label{tab:stats}
\end{table}

\begin{table}[t]
  \centering
   \begin{tabular}{|c|c|c|c|c|c|c|c|}
\hline
&\multicolumn{2}{c}{Num. Questions}&\multicolumn{3}{|c|}{Avg. Grade}&\multicolumn{2}{c|}{Num. Students} \\
  Diff.  Level&  EduRank & Asc  &  EduRank & Asc & p-value&  EduRank & Asc \\
\hline
1 & 607& 466 & 93.8  & 93.8 & 0.15&21&17\\
2 & 423& 430 & $\mathbf{94.2}$  & 90.8 & $2.8\cdot10^{-7}$ &21&17\\
3 & 449& 578 & $\mathbf{92.9}$   & 86.7 & $3.0\cdot10^{-14}$&21&17 \\
4 & 368& 233 & $\mathbf{88.5}$  & 83.0 & $3.7\cdot10^{-4}$&21&15\\
5 & 180& --- & 84.0 & --- & ---&18&-- \\
\hline
  \end{tabular}
\caption{Performance comparison between the EduRank and ASC sequencing approaches by level of difficulty (p-value: Repeated Measures ANOVA)} 

  \label{tab:per}
\end{table}
 
\section{Related Work}
\label{sec:related}
Our work relates to several areas of research in student modeling. 
 Several approaches within the educational data mining community have used computational methods for sequencing students' learning items. 
  Pardos and Heffernan~\cite{pardos2009determining} infer order over questions by predicting students' skill levels over action pairs using Bayesian knowledge tracing. They show the efficacy of this approach on a test-set comprising random sequences of three questions as well as simulated data. This approach explicitly considers each possible order sequence and does not scale to handling a large number of sequences, as in the student ranking problem we consider in this paper.

  Champaign and Cohen~\cite{champaign2010model} suggest a peer-based model for content sequencing in an intelligent tutoring system by computing the similarity between different students and choosing questions that provide the best benefit for similar students. They measure similarity by comparing between students' average performance on past questions and evaluate their approach on simulated data. Our approach differs in several ways.  First, we don't use an aggregate measure to compute similarity but compare between students' difficulty rankings over questions.  This way, we use the entire ranked list for similarity computation, and do not lose information. Consider, e.g., student1 who has accrued grades 60 and 80 on questions (a) and (b) respectively; and student2 who has accrued grades 80 and 60 on  questions (a) and (b) respectively. The average grade for both questions will be the  same despite that they clearly differ in difficulty level for the students (when ordered solely based on grade).
 Second, we are using social choice to combine similar students' difficulty ranking over questions.  Third, we evaluate our approach on two real-world data sets.  Li, Cohen and Koedinger~\cite{li2012problem} compared a blocked order approach, in which all problems of one type are completed before the student moves to the next problem type, to an interleaved approach, where problems from two types are mixed and showed that the interleaved approach yields more effective learning.  Our own approach generates an order of the different questions by reasoning about the student performance rather than determining order a-priori.

Multiple researchers have used Bayesian knowledge tracing as a way to infer students' skill acquisition (i.e., mastery level) over time given their performance levels on different question sequences ~\cite{corbett1994knowledge}.  These researchers reason about students' prior knowledge of skills and also account for slips and guessing on test problems. The models are trained on large data sets from multiple students using machine learning algorithms that account for latent variables~\cite{d2008more,falakmasir2010spectral}.  We solve a different problem --- using other students' performance to personalize ranking over test-questions. In addition, these methods measure students' performance  dichotomously (i.e., success or failure) whereas we reason about additional features such as students' grade and number of attempts to solve the question. We intend to infer students' skill levels to improve the ranking prediction in future work. 

 Approaches based on recommendation systems are increasingly being
  used in e-learning to predict students' scores and to personalize
  educational content.  We mention a few examples below and
  refer the reader to the surveys by Drachsler et al.~\cite{TEL} 
and Erdt et al.~\cite{erdt2015evaluating}
for more
  details.  Collaborative filtering (CF) was previously used in the
  educational domain for predicting students' performance. Toscher and
  Jahrer~\cite{toscher2010collaborative} use an ensemble of CF
  algorithms to predict performance for items in the KDD 2010
  educational challenge.  Berger et. al~\cite{bergner2012model} use a
  model-based approach for predicting accuracy levels of students'
  performance and skill levels on real and simulated data sets. They
  also formalize a relationship between CF and Item Response Theory
  methods and demonstrate this relationship empirically.  Schatten et al.~\cite{schatten2015integration} use matrix factorization for task sequencing in a large commercial Intelligent Tutoring System, showing improved adaptivity compared to a baseline sequencer.
  Finally, Loll
  and Pinkwart~\cite{loll2009using} use CF as a diagnostic tool for
  knowledge test questions as well as more exploratory ill-defined
  tasks. None of these approaches ranked questions according the personal difficulty level of questions to specific students. 
  
%   Furthermore, methods that predict the user grade over future questions often result in very inaccurate estimations. For example, in the KDD cup 2010, the best predictor had an average error of about 28\% of the student score, and Voss et al. \cite{voss2015transfer} that use a matrix factorization collaborative filtering algorithm also report errors above 20\% of the scores. As such, our method that directly ranks the questions by difficulty is a viable alternative to ranking by inaccurate score predictions.

\section{Discussion and Conclusion}
\label{sec:conc}
This paper presented a novel approach to personalization of educational content.  The suggested algorithm, called EduRank, combines a nearest-neighbor based collaborative filtering framework with a social choice method for preference ranking.  The algorithm constructs a
  difficulty ranking over questions for a target student by
  aggregating the ranking of similar students. 
It extends existing approaches for ranking of user items in two ways. First, 
by inferring a  difficulty
  ranking directly over the questions  for a target student, rather than
  ordering them according to predicted performance, which is prone to
  error.  Second, by penalizing disagreements between the difficulty
  rankings of similar students and the target student more highly for
   harder questions than for easy questions. 

   The algorithm was tested on two large real world data sets and its
   performance was compared to a variety of personalization methods as
   well as a non-personalized method that relied on a domain
   expert. The results showed that EduRank outperformed existing
   state-of-the-art algorithms using two metrics from the information
   retrieval literature. We extended EduRank to predict the
   difficulty rankings of questions for students with little or no
   prior history in the system, and deployed this extended version in
   a real classroom. We  show that sequencing questions using the EduRank approach in the classroom led students to
   exhibit better performance on more difficult questions, when
   compared to a baseline sequencing approach that was designed by
   domain experts.

Considering implementations in the wild, we note that EduRank was tested on two datasets with high sparsity ($>98\%$) and showed good results. Additionally, EduRank's exhibits polynomial computational complexity and was shown to be adequate for implementations in the wild, consuming $61.2$ milliseconds to compute a ranked list per student on a Mac Book air computer.

We mention several limitations of the classroom experiment and subsequent suggestions for future work. First, our classroom deployment was conducted during an elective summer course in which the usage of the e-learning system was not mandatory. As shown in Table~\ref{tab:stats} the response time for students using 
EduRank was higher than that of the students using the alternative ASC ranking approach. This may mean that students in this group were more motivated  to stay in the system.  Hence  we cannot directly  claim our results would necessarily  carry over to  actual classrooms where students exhibit a variety of motivation levels.  
  We are currently running experiments with using EduRank in a 
real classroom context in which students are assigned questions in increasing order of inferred difficulty that is outputted by EduRank.  We are also combining EduRank with a  multi-armed bandit approach to sequence 
questions to students~\cite{segal+18}.

Second, EduRank assigned more difficult questions (levels 4 and 5) than the alternative 
ASC algorithm (see Table~\ref{tab:per}), and may have adversely affected student motivation. 
In future work we mean to combine EduRank with cognitive models that directly account for skill acquisition 
and student engagement for sequencing educational content in the classroom. 

Third,    
 EduRank requires to rerun the algorithm to account for student learning over time. 
As   students solves more questions,  the change to their  ``learning state" is reflected in the change to the set  of students deemed similar to them. Consequently, running EduRank again will output a new difficulty ranking 
that is adapted to their learning state.    EduRank can be run every time it is necessary to account for effects of student learning on the inferred difficulty ranking. The complexity of running EduRank is low enough to enable EduRank to be run multiple times, depending on the needs of the teacher or education researcher. Indeed, in our study we ran EduRank the night after each class. The outputted set of questions for next class accounted for the student learning that occurred in the previous class.

%Second, the improved scores of students using EduRank in the classroom experiment could be explained by the fact they have used more time per task as opposed to resulting from a different sequencing approach. 
%We plan to conduct more experiments in the classroom to better evaluate students' behaviour and achievements with different sequencing approaches. Such experiments will control for time per task, mandatory questions and questions sequencing in separation to test the impact of each one of these factor alone and in combination.

%Second,   Edurank    does not represent or reason about students' knowledge or learning.  In future work we  intend to evaluate EduRank's performance in  comparison to such cognitive approaches.  

% Third, although we have shown that EduRank outperformed 
% the expert ranking algorithm in offline testing (Section~\ref{sec:offlineResults}) we did not compare against this method in the actual classroom. Thus we would also like to apply the expert ranking approach used in the offline study (Section 4.3) to an actual classroom.
%Finally, future work also needs to test other voting aggregation methods for EduRank's  ranking component, such as various Kemeny-Young heuristics~\cite{brandt2012computational}.

% \section{Acknowledgments}
% This work was supported in part by  EU grant no.  FP7-ICT-2011-9  \#600854
% and by ISF grant no. 1276/12.

\bibliographystyle{elsarticle-num}

\bibliography{eduRefs}

\begin{thebibliography}{10}
\expandafter\ifx\csname url\endcsname\relax
  \def\url#1{\texttt{#1}}\fi
\expandafter\ifx\csname urlprefix\endcsname\relax\def\urlprefix{URL }\fi
\expandafter\ifx\csname href\endcsname\relax
  \def\href#1#2{#2} \def\path#1{#1}\fi

\bibitem{sampson2010personalised}
D.~Sampson, C.~Karagiannidis, Personalised learning: Educational, technological
  and standardisation perspective, Interactive Educational Multimedia 4 (2002)
  24--39.

\bibitem{akbulut2012adaptive}
Y.~Akbulut, C.~S. Cardak, Adaptive educational hypermedia accommodating
  learning styles: A content analysis of publications from 2000 to 2011,
  Computers \& Education 58~(2) (2012) 835--842.

\bibitem{ba2007framework}
H.~Ba-Omar, I.~Petrounias, F.~Anwar, A framework for using web usage mining to
  personalise e-learning, in: Advanced Learning Technologies, 2007. ICALT 2007.
  Seventh IEEE International Conference on, IEEE, 2007, pp. 937--938.

\bibitem{zhang2008personalized}
L.~Zhang, X.~Liu, X.~Liu, Personalized instructing recommendation system based
  on web mining, in: Young Computer Scientists, 2008. ICYCS 2008. The 9th
  International Conference for, IEEE, 2008, pp. 2517--2521.

\bibitem{najar2014adaptive}
A.~S. Najar, A.~Mitrovic, B.~M. McLaren, Adaptive support versus alternating
  worked examples and tutored problems: Which leads to better learning?, in:
  International Conference on User Modeling, Adaptation, and Personalization,
  Springer, 2014, pp. 171--182.

\bibitem{mazziotti2015robust}
C.~Mazziotti, W.~Holmes, M.~Wiedmann, K.~Loibl, N.~Rummel, M.~Mavrikis,
  A.~Hansen, B.~Grawemeyer, Robust student knowledge: Adapting to individual
  student needs as they explore the concepts and practice the procedures of
  fractions, in: Workshop on Intelligent Support in Exploratory and Open-Ended
  Learning Environments Learning Analytics for Project Based and Experiential
  Learning Scenarios at the 17th International Conference on Artificial
  Intelligence in Education (AIED 2015), 2015, pp. 32--40.

\bibitem{breese1998empirical}
J.~S. Breese, D.~Heckerman, C.~Kadie, Empirical analysis of predictive
  algorithms for collaborative filtering, in: Proceedings of the Fourteenth
  conference on Uncertainty in artificial intelligence, Morgan Kaufmann
  Publishers Inc., 1998, pp. 43--52.

\bibitem{shani2011evaluating}
G.~Shani, A.~Gunawardana, Evaluating recommendation systems, in: Recommender
  systems handbook, Springer, 2011, pp. 257--297.

\bibitem{toscher2010collaborative}
A.~Toscher, M.~Jahrer, Collaborative filtering applied to educational data
  mining, KDD Cup.

\bibitem{liu2008eigenrank}
N.~N. Liu, Q.~Yang, Eigenrank: a ranking-oriented approach to collaborative
  filtering, in: Proceedings of the 31st annual international ACM SIGIR
  conference on Research and development in information retrieval, ACM, 2008,
  pp. 83--90.

\bibitem{brandt2012computational}
F.~Brandt, V.~Conitzer, U.~Endriss, Computational social choice, Multiagent
  systems (2012) 213--283.

\bibitem{EduRank}
A.~Segal, Z.~Katzir, K.~Gal, G.~Shani, B.~Shapira, Edurank: A collaborative
  filtering approach to personalization in e-learning, in: Educational Data
  Mining 2014, 2014.

\bibitem{graesser2012intelligent}
A.~C. Graesser, M.~W. Conley, A.~Olney, Intelligent tutoring systems, APA
  handbook of educational psychology. Washington, DC: American Psychological
  Association.

\bibitem{kizilcec2017towards}
R.~F. Kizilcec, G.~M. Davis, G.~L. Cohen, Towards equal opportunities in moocs:
  affirmation reduces gender \& social-class achievement gaps in china, in:
  Proceedings of the Fourth (2017) ACM Conference on Learning@ Scale, ACM,
  2017, pp. 121--130.

\bibitem{clow2013moocs}
D.~Clow, Moocs and the funnel of participation, in: Proceedings of the Third
  International Conference on Learning Analytics and Knowledge, ACM, 2013, pp.
  185--189.

\bibitem{zhao2006shortest}
C.~Zhao, L.~Wan, A shortest learning path selection algorithm in e-learning,
  in: Advanced Learning Technologies, 2006. Sixth International Conference on,
  IEEE, 2006, pp. 94--95.

\bibitem{idris2009adaptive}
N.~Idris, N.~Yusof, P.~Saad, Adaptive course sequencing for personalization of
  learning path using neural network, International Journal of Advances in Soft
  Computing and its Applications 1~(1) (2009) 49--61.

\bibitem{al2011heuristic}
S.~A. Al-Radaei, R.~Mishra, A heuristic method for learning path sequencing for
  intelligent tutoring system (its) in e-learning, International Journal of
  Intelligent Information Technologies (IJIIT) 7~(4) (2011) 65--80.

\bibitem{sarwar}
B.~Sarwar, G.~Karypis, J.~Konstan, J.~Riedl, Application of dimensionality
  reduction in recommender system-a case study, Tech. rep., DTIC Document
  (2000).

\bibitem{KorenB15}
Y.~Koren, R.~M. Bell, Advances in collaborative filtering, in: Recommender
  Systems Handbook, 2015, pp. 77--118.

\bibitem{weimer2008cofi}
M.~Weimer, A.~Karatzoglou, Q.~V. Le, A.~J. Smola, Cofi rank-maximum margin
  matrix factorization for collaborative ranking, in: Advances in neural
  information processing systems, 2008, pp. 1593--1600.

\bibitem{fishburn1973theory}
P.~C. Fishburn, The theory of social choice, Vol. 264, Princeton University
  Press Princeton, 1973.

\bibitem{copeland1951reasonable}
A.~H. Copeland, A reasonable social welfare function, in: University of
  Michigan Seminar on Applications of Mathematics to the social sciences, 1951.

\bibitem{nurmi1983voting}
H.~Nurmi, Voting procedures: a summary analysis, British Journal of Political
  Science 13~(02) (1983) 181--208.

\bibitem{schalekamp2009rank}
F.~Schalekamp, A.~van Zuylen, Rank aggregation: Together we're strong., in:
  ALENEX, 2009, pp. 38--51.

\bibitem{pennock2000social}
D.~M. Pennock, E.~Horvitz, C.~L. Giles, et~al., Social choice theory and
  recommender systems: Analysis of the axiomatic foundations of collaborative
  filtering, in: AAAI/IAAI, 2000, pp. 729--734.

\bibitem{zhou2010evaluating}
B.~Zhou, Y.~Yao, Evaluating information retrieval system performance based on
  user preference, Journal of Intelligent Information Systems 34~(3) (2010)
  227--248.

\bibitem{yao1995measuring}
Y.~Yao, Measuring retrieval effectiveness based on user preference of
  documents, JASIS 46~(2) (1995) 133--145.

\bibitem{shanievaluating}
G.~Shani, A.~Gunawardana, Evaluating recommendation systems, in: F.~Ricci,
  L.~Rokach, B.~Shapira, P.~B. Kantor (Eds.), Recommender Systems Handbook,
  Springer US, 2011, pp. 257--297.

\bibitem{yilmaz2008new}
E.~Yilmaz, J.~A. Aslam, S.~Robertson, A new rank correlation coefficient for
  information retrieval, in: Proceedings of the 31st annual international ACM
  SIGIR conference on Research and development in information retrieval, ACM,
  2008, pp. 587--594.

\bibitem{koedinger2010data}
K.~R. Koedinger, R.~Baker, K.~Cunningham, A.~Skogsholm, B.~Leber, J.~Stamper, A
  data repository for the edm community: The pslc datashop, Handbook of
  educational data mining (2010) 43--55.

\bibitem{wauters2012item}
K.~Wauters, P.~Desmet, W.~Van Den~Noortgate, Item difficulty estimation: An
  auspicious collaboration between data and judgment, Computers \& Education
  58~(4) (2012) 1183--1193.

\bibitem{pardos2011kt}
Z.~A. Pardos, N.~T. Heffernan, Kt-idem: Introducing item difficulty to the
  knowledge tracing model, in: User Modeling, Adaption and Personalization,
  Springer, 2011, pp. 243--254.

\bibitem{baker2008students}
R.~S. Baker, J.~Walonoski, N.~Heffernan, I.~Roll, A.~Corbett, K.~Koedinger, Why
  students engage in gaming the system behavior in interactive learning
  environments, Journal of Interactive Learning Research 19~(2) (2008)
  185--224.

\bibitem{kanoulas2009empirical}
E.~Kanoulas, J.~A. Aslam, Empirical justification of the gain and discount
  function for ndcg, in: Proceedings of the 18th ACM conference on Information
  and knowledge management, ACM, 2009, pp. 611--620.

\bibitem{zhou2008large}
Y.~Zhou, D.~Wilkinson, R.~Schreiber, R.~Pan, Large-scale parallel collaborative
  filtering for the netflix prize, in: Algorithmic Aspects in Information and
  Management, Springer, 2008, pp. 337--348.

\bibitem{schatten2015integration}
C.~Schatten, R.~Janning, L.~Schmidt-Thieme, Integration and evaluation of a
  matrix factorization sequencer in large commercial its., in: AAAI, 2015, pp.
  1380--1386.

\bibitem{schelter2012collaborative}
S.~Schelter, S.~Owen, Collaborative filtering with apache mahout, Proc. of ACM
  RecSys Challenge.

\bibitem{corbett1994knowledge}
A.~T. Corbett, J.~R. Anderson, Knowledge tracing: Modeling the acquisition of
  procedural knowledge, User modeling and user-adapted interaction 4~(4) (1994)
  253--278.

\bibitem{Schein:2002:MMC:564376.564421}
A.~I. Schein, A.~Popescul, L.~H. Ungar, D.~M. Pennock, Methods and metrics for
  cold-start recommendations, in: Proceedings of the 25th Annual International
  ACM SIGIR Conference on Research and Development in Information Retrieval,
  SIGIR '02, ACM, New York, NY, USA, 2002, pp. 253--260.
\newblock \href {http://dx.doi.org/10.1145/564376.564421}
  {\path{doi:10.1145/564376.564421}}.

\bibitem{block1971mastery}
J.~H. Block, P.~W. Airasian, B.~S. Bloom, J.~B. Carroll, Mastery learning:
  Theory and practice, Holt, Rinehart and Winston New York, 1971.

\bibitem{pardos2009determining}
Z.~A. Pardos, N.~T. Heffernan, Determining the significance of item order in
  randomized problem sets., in: EDM, 2009.

\bibitem{champaign2010model}
J.~Champaign, R.~Cohen, A model for content sequencing in intelligent tutoring
  systems based on the ecological approach and its validation through simulated
  students., in: FLAIRS Conference, 2010.

\bibitem{li2012problem}
N.~Li, W.~W. Cohen, K.~R. Koedinger, Problem order implications for learning
  transfer, in: Intelligent Tutoring Systems, Springer, 2012, pp. 185--194.

\bibitem{d2008more}
R.~S. Baker, A.~T. Corbett, V.~Aleven, More accurate student modeling through
  contextual estimation of slip and guess probabilities in bayesian knowledge
  tracing, in: Intelligent Tutoring Systems, Springer, 2008, pp. 406--415.

\bibitem{falakmasir2010spectral}
M.~H. Falakmasir, Z.~A. Pardos, G.~J. Gordon, P.~Brusilovsky, A spectral
  learning approach to knowledge tracing, in: EDM, 2013.

\bibitem{TEL}
H.~Drachsler, K.~Verbert, O.~Santos, N.~Manouselis, Panorama of recommender
  systems to support learning, in: F.~Ricci, L.~Rokach, B.~Shapira (Eds.),
  Recommender Systems Handbook, Springer US, 2015, pp. 421--451.

\bibitem{erdt2015evaluating}
M.~Erdt, A.~Fernandez, C.~Rensing, Evaluating recommender systems for
  technology enhanced learning: A quantitative survey, Learning Technologies,
  IEEE Transactions on 8~(4) (2015) 326--344.

\bibitem{bergner2012model}
Y.~Bergner, S.~Droschler, G.~Kortemeyer, S.~Rayyan, D.~Seaton, D.~Pritchard,
  Model-based collaborative filtering analysis of student response data:
  Machine-learning item response theory., in: EDM, 2012, pp. 95--102.

\bibitem{loll2009using}
F.~Loll, N.~Pinkwart, Using collaborative filtering algorithms as elearning
  tools, in: 42nd Hawaii International Conference on Systems Science, 2009.

\bibitem{segal+18}
A.~Segal, Y.~Ben~David, J.~J. Williams, K.~Gal, Y.~Shalom, Combining difficulty
  ranking with multi-armed bandits to sequence educational content, in:
  C.~Penstein~Ros{\'e}, R.~Mart{\'i}nez-Maldonado, H.~U. Hoppe, R.~Luckin,
  M.~Mavrikis, K.~Porayska-Pomsta, B.~McLaren, B.~du~Boulay (Eds.), Artificial
  Intelligence in Education, Springer International Publishing, Cham, 2018, pp.
  317--321.

\end{thebibliography}

\end{document}